\documentclass[journal]{IEEEtran}
\usepackage[boxed]{algorithm2e}
\usepackage{subfigure, graphicx}
\usepackage{amsmath,amssymb}
\usepackage{multirow}
\usepackage{epstopdf}

\newtheorem{theorem}{Theorem}
\newtheorem{lemma}{Lemma}

\ifCLASSINFOpdf
\else
\fi
\begin{document}
%
\title{An Iteratively Re-weighted Method for Problems with Sparsity-Inducing Norms}
%
%
%
\author{Feiping Nie, Zhanxuan Hu, Xiaoqian Wang,
 Rong Wang,
        Xuelong Li,~\IEEEmembership{Fellow,~IEEE}, Heng Huang
\IEEEcompsocitemizethanks{\IEEEcompsocthanksitem F. Nie, Z. Hu, R. Wang and X. Li are with the School
of Computer Science, OPTIMAL, Northwestern Polytechnical University, Xian 710072, Shaanxi, P. R. China.
E-mail: {huzhanxuan@mail.nwpu.edu.cn, feipingnie@gmail.com, wangrong07@tsinghua.org.cn, li@nwpu.edu.cn}
\IEEEcompsocthanksitem X. Wang and H. Huang are with Computer Engineering,  University of Pittsburgh,  Pittsburgh, PA 15261, USA. E-mail: {joy.xqwang@gmail.com, heng.huang@pitt.edu}}
}

%
%

\markboth{Journal of \LaTeX\ Class Files,~Vol.~14, No.~8, August~2015}%
{Shell \MakeLowercase{\textit{et al.}}: Bare Demo of IEEEtran.cls for IEEE Journals}
%



\maketitle

\begin{abstract}
This work aims at solving the problems with intractable sparsity-inducing norms that are often encountered in various machine learning tasks, such as multi-task learning, subspace clustering, feature selection, robust principal component analysis, and so on. Specifically, an Iteratively Re-Weighted method~(IRW) with solid convergence guarantee is provided. We investigate its convergence speed via numerous experiments on real data. Furthermore, in order to validate the practicality of IRW, we use it to solve a concrete robust feature selection model with complicated objective function. The experimental results show that the model coupled with proposed optimization method outperforms alternative methods significantly.
\end{abstract}


%
\IEEEpeerreviewmaketitle
\section{Introduction}

The problem of regularized risk minimization is encountered in many machine learning fields. It aims to admit a tradeoff between regularizer and loss function as:
\begin{equation}\label{eq:1}
  \min_{x\in \mathcal{C}} f(x)+\mu g(x)\,,
\end{equation}
where $f(x)$ denotes the loss function, $g(x)$ denotes the regularizer, and $\mu$ is a regularization parameter balancing these two terms. Generally, the loss function $f(x)$ is relevant to the problem we aim to solve, while the regularizer $g(x)$ depends on the assumption over the structure of $x$. In practice,  the sparsity-inducing norms is a representative example for both $f(x)$ and $g(x)$, and has been widely used to cope with various machine learning tasks, such as feature selection~\cite{nie2010efficient}, subspace clustering~\cite{elhamifar2009sparse,liu2013robust}, multi-task learning~\cite{argyriou2008convex}. We provide a short summarization for representative models in Table~\ref{tab:m1}, where the notations of $\|\bullet\|_0$ and $\|\bullet\|_{2,0}$ can be found in Section~\ref{sec:1}, and the details over these tasks can be found in the original papers. In addition, note that in this paper we consider $rank(X)$ as a sparsity-inducing norm, as $rank(X)=\sum_{i=1}(\sigma_i(X))^0$, where $\{\sigma_i(X)\}$ are the singular values of $X$.

Solving the models with sparsity-inducing norms mentioned in Table~\ref{tab:m1} is generally an NP-hard problem, and a general method is looking for
some relaxations that solve the original objective functions approximately but efficiently. Most of the relaxations are convex, such as the $\ell_1$-norm, $\ell_{2,1}$-norm, and nuclear norm. In addition to solid theoretical guarantees, a significant advantage of using convex relaxations is that the involved problems can be solved efficiently by some traditional optimization methods, including Alternating Direction Method of Multipliers~(ADMM), Frank-Wolfe (FW) algorithm, proximal algorithm, and stochastic gradient descent~(SGD)~\cite{yao2017efficient}.

In practice, however, convex relaxations often lead to an over-penalized problem~\cite{gong2013general}. To alleviate this issue, numerous non-convex relaxations have been proposed, such as $\ell_p$~(Schatten p)-norm~\cite{mohan2012iterative}, Capped-$\ell_1$-norm~\cite{nie2017joint}, Truncated Nuclear Norm~\cite{zhang2012matrix}, MCP~\cite{zhang2010nearly}, SCAD~\cite{fan2001variable}. Although the non-convex relaxations have achieved great success in several practical applications, how to solve the involved problems is still challenging.  Concave-Convex Procedure~(CCP)~\cite{yuille2002concave} is a principled approach to tackle the non-convex problems. Nevertheless, its practicability is generally limited by the high time cost in solving subproblem. Recently, numerous efforts have been made to generalize the proximal algorithm to non-convex problems, such as general iterative shrinkage and thresholding~(GIST)~\cite{gong2013general},  Inertial Forward-Backward~(IFB)~\cite{boct2016inertial}, nonmonotone Accelerated Proximal Gradient~(nonAPG)~\cite{li2015accelerated}, and Redistributing Nonconvexity~\cite{yao2017efficient}. But, none of them can tackle the case that both loss function and regularizer are non-convex and non-smooth.
\begin{table}[]
\centering
\caption{Machine learning tasks with Sparsity-Inducing Norms}\label{tab:m1}
\begin{tabular}{|c|c|}
\hline
Task                                 & Model \\ \hline \hline
\multirow{2}{*}{Subspace clustering~\cite{elhamifar2009sparse,liu2013robust}} &   $\min_{W\in \mathcal{C}} \|X-XW\|_0+\mu\|W\|_0$     \\ \cline{2-2}
                                     &   $\min_{W\in \mathcal{C}} \|X-XW\|_0+\mu rank(W)$    \\ \hline
\multirow{2}{*}{Multi-task Learning~\cite{argyriou2008convex,pong2010trace}} &   $\min_{W} \sum_{i=1}^{K}\|X_i^Tw_i-y_i\|_2^2 + \mu \|W\|_{2,0} $    \\ \cline{2-2}
                                     &  $\min_{W} \sum_{i=1}^{K}\|X_i^Tw_i-y_i\|_2^2 + \mu rank(W) $      \\ \hline
RPCA~\cite{lin2010augmented}                                 &  $\min_{W\in \mathcal{C}} \|X-W\|_0+\mu rank(W)$     \\ \hline
\multirow{1}{*}{Matrix Completion~\cite{candes2009exact}}   &  $\min_{W\in \mathcal{C}} \|P_\Omega(X-W)\|_0+\mu rank(W)$      \\ \hline
\multirow{1}{*}{Feature Selection~\cite{nie2010efficient}}   &  $\min_{W}\|X^TW-Y\|_{2,0}+\mu\|W\|_{2,0}$      \\ \hline
\end{tabular}
\end{table}
The Iteratively Re-Weighted Method~(IRW) that we focus in this work has been used in previous studies~\cite{nie2012low,nie2018calibrated,lu2014generalized,mohan2012iterative,candes2008enhancing,sun2017majorization}, but all of them aim at solving the model involving only the low rank regularizers. In this work, we generalize IRW to a general problem, where the objective function has multiple Sparsity-Inducing Norms including Schatten p-norm, i.e., the low rank regularizer~\footnote{Note that both $\|\bullet\|_0$, $\|\bullet\|_{2,0}$, $rank(\bullet)$ and their relaxations are named as sparsity-inducing norms, but in this paper we mainly focus on the latter.}. The key principle of IRW lies in finding a surrogate function with the following two properties:
\begin{itemize}
  \item Convexity and smoothness;
  \item The closed-form solution can be solved efficiently.
\end{itemize}
We show that the original complicated problem can be solved efficiently via iteratively solving the surrogate function. In addition,  we provide a solid theoretical analysis for proposed method, and conduct numerous experiments on real data to investigate its convergence speed. In order to further validate the practicality of  proposed method, we utilize it to cope with a novel robust feature selection model developed in this paper. Numerous experimental results demonstrate that the model coupled with the proposed optimization method IRW provides a large advantage over alternative algorithms.

\section{Notations and Definitions}\label{sec:1}
\vspace*{-3pt}
In this paper, we use lowercase letter if it would be scalar, vector or matrix, and use uppercase letter if it is matrix.
For matrix $M$, its $i$-th row, $j$-th column and the $ij$-th entry of ${M}$
are denoted by ${m}^i$, ${m}_j$ and $m_{ij}$, respectively. $tr(M)$ denotes the trace of matrix $M$.

The $\ell_{r,p}$-norm of matrix $M$ is defined as
\begin{equation}
\label{rpnorm}
\left\| {M} \right\|_{r,p}  = \left( {\sum\limits_{i = 1}^n {\left( {\sum\limits_{j = 1}^m {\left|m_{ij}\right|^r } } \right)^{\frac{p}{r}} } } \right)^{\frac{1}{p}}  = \left( {\sum\limits_{i = 1}^n {\left\| {{m}^i } \right\|_r^p } } \right)^{\frac{1}{p}}.
\end{equation}
Particularly, when $r\ge 1$ and $p\ge 1$, $\ell_{r,p}$-norm is a valid {\it norm} because it
satisfies the three norm conditions, including the triangle
inequality $\|{A}\|_{r,p} +\|{B}\|_{r,p} \geq
\|{A+B}\|_{r,p}$. When $r< 1$ or $p< 1$, the $\ell_{r,p}$-norm is not a valid {\it norm}, the term ``{\em norm}" here is for convenience. In Eq.(\ref{rpnorm}), when $M$ becomes a column or row vector $m$, the $\ell_{r,p}$-norm of $M$ is reduced to the $\ell_{p}$-norm of $m$.

The Schatten $p$-norm of a matrix ${M}$ was defined as
\begin{equation}
\left\| {M} \right\|_{S_p}  = \left( \sum\limits_{i = 1} {\sigma_{i}^p } \right)^\frac{1}{p}
= \left(tr(({M}^T{M})^{\frac{p}{2}}) \right)^\frac{1}{p},
\end{equation}
where $\sigma_{i}$ is the $i$-th singular value of ${M}$. 
When $p\ge 1$, Schatten $p$-norm is a valid {\it norm}. When $p< 1$, the Schatten $p$-norm is not a valid {\it norm}, the term ``{\em norm}" here is for convenience.

\section{Iteratively Reweighted Method for A General Sparse Coding Problem}

\subsection{A General Sparse Coding Problem}

In this section, we focus on solving a general problem as follows:
\begin{equation}
\label{opgp}
\mathop {\min }\limits_{x \in \mathcal{C}} f(x) + \mu\sum\limits_i {tr((g_i^T (x)g_i (x))^{\frac{p}{2}} )}\,.
\end{equation}
Note that when $g_i (x)$ is scalar, vector or matrix output function, then $tr((g_i^T (x)g_i (x))^{\frac{p}{2}} )$ becomes the following terms respectively:
\begin{equation}
tr((g_i^T (x)g_i (x))^{\frac{p}{2}} ) = \left\{ {\begin{array}{*{20}c}
   {\left| {g_i (x)} \right|^p } & {g_i (x)\;is\;scalar}  \\
   {\left\| {g_i (x)} \right\|_2^p } & {g_i (x)\;is\;vector}  \\
   {\left\| {g_i (x)} \right\|_{S_p }^p } & {g_i (x)\;is\;matrix}  \\
\end{array}} \right. \,.
\end{equation}
For the case that $p=1$, $tr((g_i^T (x)g_i (x))^{\frac{p}{2}} )$ denotes the $\ell_{1}$-norm, $\ell_{2}$-norm and trace norm respectively,
\begin{equation}
tr((g_i^T (x)g_i (x))^{\frac{1}{2}} ) = \left\{ {\begin{array}{*{20}c}
   {\left| {g_i (x)} \right|} & {g_i (x)\;is\;scalar}  \\
   {\left\| {g_i (x)} \right\|_2 } & {g_i (x)\;is\;vector}  \\
   {\left\| {g_i (x)} \right\|_* } & {g_i (x)\;is\;matrix}  \\
\end{array}} \right.
\end{equation}
For the Eq. (\ref{opgp}) is non-smooth, we can turn to solve a approximation problem of it, a smooth problem formulated as follows:
\begin{equation}
\label{opgp1}
\mathop {\min }\limits_{x \in C} f(x) + \mu\sum\limits_i {tr((g_i^T (x)g_i (x) + \delta I)^{\frac{p}{2}} )}\,.
\end{equation}
And, when $\delta  \to 0$, Eq.(\ref{opgp1}) is reduced to Eq.(\ref{opgp}) since the following equations hold:
\begin{equation}
\mathop {\lim }\limits_{\delta  \to 0} tr((g_i^T (x)g_i (x) + \delta I)^{\frac{p}{2}} ) = \left\{ {\begin{array}{*{20}c}
   {\left| {g_i (x)} \right|^p } & {g_i (x)\;is\;scalar}  \\
   {\left\| {g_i (x)} \right\|_2^p } & {g_i (x)\;is\;vector}  \\
   {\left\| {g_i (x)} \right\|_{S_p }^p } & {g_i (x)\;is\;matrix}  \\
\end{array}} \right.\,.
\end{equation}

Next, we focus on solving the approximation problem~\eqref{opgp1}.

\section{Iteratively Reweighted Algorithm for the Approximation Problem}
Before deriving the algorithm for optimizing the problem \eqref{opgp1}, we need some significant lemmas as follows.
First, according to the chain rule in calculus, we have Lemma.1.
\begin{lemma}[Chain rule]
\label{lem1}
Suppose $g(x)$ is a matrix output function, $h(x)$ is a scalar output function, $x$ is a scalar, vector or matrix variable, then we have
\begin{equation}
\begin{split}
     & \frac{{\partial h(g(x))}}{{\partial x}} = \frac{{\sum\limits_{i,j} {\frac{{\partial h(g(x))}}{{\partial g_{ij} (x)}}\partial g_{ij} (x)} }}{{\partial x}}  \\
     & = \frac{{tr\left( {\left( {\frac{{\partial h(g(x))}}{{\partial g(x)}}} \right)^T \partial g(x)} \right)}}{{\partial x}}
\end{split}
\end{equation}
\end{lemma}
According to the chain rule in Lemma~\ref{lem1}, we have the following two lemmas:
\begin{lemma}
\label{lem2}
Suppose $g (x)$ is a scalar, vector or matrix output function, $x$ is a scalar, vector or matrix variable, then we have
\begin{equation}\label{cr1}
\begin{split}
     & \frac{{\partial tr((g^T (x)g(x) + \delta I)^{\frac{p}{2}} )}}{{\partial x}} \\
     &  = \frac{{tr\left( {2\frac{p}{2}(g^T (x)g(x) + \delta I)^{\frac{{p - 2}}{2}} g^T (x)\partial g(x)} \right)}}{{\partial x}}\,.
\end{split}
\end{equation}
\end{lemma}
\textbf{Proof.}
Let $h(x) = tr(x^Tx + \delta I)^{\frac{p}{2}} $, we have
\begin{equation}
\frac{{\partial h(x)}}{{\partial x}} = 2\frac{p}{2}x(x^T x + \delta I)^{\frac{{p - 2}}{2}}\,,
\end{equation}
further, we can obtain
 \begin{equation}
 {\frac{{\partial h(g(x))}}{{\partial g(x)}}} = {2\frac{p}{2}g(x)(g^T (x)g(x) + \delta I)^{\frac{{p - 2}}{2}} }\,.
 \end{equation}
According to the chain rule in Lemma \ref{lem1}, we get the Eq.(\ref{cr1}).
\hfill $\Box$

\begin{lemma}
\label{lem3}
Suppose $g (x)$ is a scalar, vector or matrix output function, $x$ is a scalar, vector or matrix variable, $D$ is a constant and $D$ is symmetrical if $D$ is a matrix, then we have
\begin{equation}
\label{cr2}
\frac{{\partial tr(g^T (x)g(x)D)}}{{\partial x}} = \frac{{tr\left( {2Dg^T (x)\partial g(x))} \right)}}{{\partial x}}\,.
\end{equation}
\end{lemma}
\textbf{Proof.}
Let $h(x) = tr(x^TxD)$, we have $\frac{{\partial h(x)}}{{\partial x}} = 2xD$, then we have ${\frac{{\partial h(g(x))}}{{\partial g(x)}}} = 2g(x)D$. So according to the chain rule in Lemma \ref{lem1}, we get the Eq.(\ref{cr2}).
\hfill $\Box$

\subsection{Algorithm Derivation}
Now we derive the algorithm for optimizing the problem (\ref{opgp1}). The Lagrangian function of the problem (\ref{opgp1}) is
\begin{equation}
\label{lag1}
  \mathcal{L}(x,\lambda ) = f(x) + \mu \sum\limits_i {tr((g_i^T (x)g_i (x) + \delta I)^{\frac{p}{2}} )}  - r(x,\lambda ),
\end{equation}
where $r(x,\lambda )$ is a Lagrangian term for the constraint $x \in \mathcal{C}$. By setting the derivative of Eq.(\ref{lag1}) w.r.t. $x$ to zero, we have
\begin{equation}\label{kkt0}
\begin{split}
      \frac{{\partial L(x,\lambda )}}{{\partial x}}&= f'(x) +  \mu \sum\limits_i {\frac{{\partial tr((g_i^T (x)g_i (x) + \delta I)^{\frac{p}{2}} )}}{{\partial x}}} \\
     &   - \frac{{\partial r(x,\lambda )}}{{\partial x}} = 0\,.
\end{split}
\end{equation}
According to Lemma \ref{lem2}, Eq.(\ref{kkt0}) can be rewritten as
\begin{equation}\label{kkt}
\begin{split}
     & f'(x) + \mu\sum\limits_i {\frac{{tr\left( {2\frac{p}{2}(g_i^T (x)g_i (x) + \delta I)^{\frac{{p - 2}}{2}} g_i^T (x)\partial g_i (x)} \right)}}{{\partial x}}} \\
     &   - \frac{{\partial r(x,\lambda )}}{{\partial x}} = 0\,.
\end{split}
\end{equation}
If we can find a solution $x$ that satisfies the Eq.(\ref{kkt}), then we usually find a stationary point or global optimal solution to the problem (\ref{opgp1}) according to the
Karush-Kuhn-Tucker conditions. However, directly finding a solution $x$ that satisfies Eq.(\ref{kkt}) is generally not an easy task. In this paper, we propose an iterative
algorithm to find it. A basic observation is that, if $D_i  = \frac{p}{2}(g_i^T (x)g_i (x) + \delta I)^{\frac{{p - 2}}{2}}$ is a given constant, then Eq.(\ref{kkt})
is reduced to
\begin{equation}
\label{kkteasy}
f'(x) + \mu\sum\limits_i {\frac{{tr\left( {2D_i g_i^T (x)\partial g_i (x)} \right)}}{{\partial x}}}  - \frac{{\partial r(x,\lambda )}}{{\partial x}} = 0\,.
\end{equation}
which is equivalent to solving the following problem:
\begin{equation}
\label{opeasy}
\mathop {\min }\limits_{x \in \mathcal{C}} f(x) + \mu\sum\limits_i {tr(g_i^T (x)g_i (x)D_i )}\,.
\end{equation}
Based on the observation, we first guess a solution $x$, then we calculate $D_i$ based on the current solution $x$ and then update the current solution $x$ by the optimal solution of the problem (\ref{opeasy}) based on the calculated $D_i$. We iteratively perform this procedure until it converges.
The detailed algorithm is described in Algorithm~\ref{alg1}. We will give a theoretical analysis to prove the convergence of the proposed algorithm.
\begin{algorithm}
\label{alg1}
Initialize $x \in \mathcal{C}$  \;
\While{not converge}
{ 1. For each $i$, calculate $D_i  = \frac{p}{2}(g_i^T (x)g_i (x) + \delta I)^{\frac{{p - 2}}{2}}$  \;
2. Update $x$ by solving the problem $\mathop {\min }\limits_{x \in \mathcal{C}} f(x) + \mu\sum\limits_i {tr(g_i^T (x)g_i (x)D_i )}$ \;
}
\KwOut{$x$.}
\caption{The algorithm to solve the problem~(\ref{opgp1}).}
\end{algorithm}

\subsection{Convergence Analysis of Algorithm \ref{alg1}}
Before proving the convergence of the Algorithm~\ref{alg1}, we first introduce several significant lemmas.
\begin{lemma}
\label{lemma1}
For any $\sigma  > 0$, the following inequality holds when $0<p \le 2$:
\begin{equation}
\label{ineq01}
\frac{p}{2}\sigma  - \sigma ^{\frac{p}{2}}  + \frac{{2 - p}}{2} \ge 0\,.
\end{equation}
\end{lemma}
\textbf{Proof.} Denote $f(\sigma)=p\sigma  - 2\sigma ^{\frac{p}{2}}  + 2 - p$,
we have the following derivatives:
\begin{equation}
f'(\sigma)=p(1-\sigma ^{\frac{p-2}{2}}),
\quad
\textnormal{and}
\quad
f''(\sigma)=\frac{p(2-p)}{2}\sigma_i ^{\frac{p-4}{2}}. \nonumber
\end{equation}
Obviously, when $\sigma > 0$ and $0<p \le 2$, then $f''(\sigma) \ge 0$ and $\sigma=1$ is the only point that $f'(\sigma)=0$. Note that $f(1)=0$, thus when $\sigma > 0$ and $0<p \le 2$, then $f(\sigma) \ge 0$, which indicates Eq.(\ref{ineq01}).
\hfill $\Box$

\begin{lemma}[\cite{traceineq70}]
\label{lemma2}
For any positive definite matrices $\tilde M,M$ with the same size, suppose the eigen-decomposition $\tilde M=U\Sigma U^T$, $M=V\Lambda V^T$, where
the eigenvalues in $\Sigma$ is in increasing order and the eigenvalues in $\Lambda$ is in decreasing order. Then the following inequality holds:
\begin{equation}
\label{ineq2}
tr(\tilde MM) \ge tr(\Sigma \Lambda )\,.
\end{equation}
\end{lemma}
\begin{lemma}
\label{lemma3}
For any positive definite matrices $\tilde M,M$ with the same size, the following inequality holds when $0<p \le 2$.
\begin{equation}
\label{ineq3}
tr(\tilde M^{\frac{p}{2}} ) - \frac{p}{2}tr(\tilde MM^{\frac{{p - 2}}{2}} ) \le tr(M^{\frac{p}{2}} ) - \frac{p}{2}tr(MM^{\frac{{p - 2}}{2}} )\,.
\end{equation}
\end{lemma}
\textbf{Proof.}
For any $\sigma  > 0$, $\lambda  > 0$ and $0<p \le 2$, according to Lemma \ref{lemma1} we have
$\frac{p}{2}(\frac{\sigma }{\lambda }) - (\frac{\sigma }{\lambda })^{\frac{p}{2}}  + \frac{{2 - p}}{2} \ge 0$, which indicates
\begin{equation}
\label{ineq31}
 \frac{p}{2}\sigma \lambda ^{\frac{{p - 2}}{2}}  - \sigma ^{\frac{p}{2}}  + \frac{{2 - p}}{2}\lambda ^{\frac{p}{2}}  \ge 0\,.
\end{equation}
Suppose the eigen-decomposition $\tilde M=U\Sigma U^T$, $M=V\Lambda V^T$, where
the eigenvalues in $\Sigma$ is in increasing order and the eigenvalues in $\Lambda$ is in decreasing order. Then, according to Eq.(\ref{ineq31}),
we have
\begin{equation}
\frac{p}{2}tr(\Sigma \Lambda ^{\frac{{p - 2}}{2}} ) - tr(\Sigma ^{\frac{p}{2}} ) + \frac{{2 - p}}{2}tr(\Lambda ^{\frac{p}{2}} ) \ge 0,
\end{equation}
and according to Lemma \ref{lemma2} we have
\begin{equation}
\frac{p}{2}tr(\tilde MM^{\frac{{p - 2}}{2}} ) - \frac{p}{2}tr(\Sigma \Lambda ^{\frac{{p - 2}}{2}} ) \ge 0\,.
\end{equation}
and
\begin{equation}
\frac{p}{2}tr(\tilde MM^{\frac{{p - 2}}{2}} ) - tr(\Sigma ^{\frac{p}{2}} ) + \frac{{2 - p}}{2}tr(\Lambda ^{\frac{p}{2}} ) \ge 0\,.
\end{equation}
Note that $tr(\tilde M^{\frac{p}{2}} ) = tr(\Sigma ^{\frac{p}{2}} )$ and $tr(M^{\frac{p}{2}} ) = tr(\Lambda ^{\frac{p}{2}} )$, so we have
\[
\begin{array}{l}
 \frac{p}{2}tr(\tilde MM^{\frac{{p - 2}}{2}} ) - tr(\tilde M^{\frac{p}{2}} ) + \frac{{2 - p}}{2}tr(M^{\frac{p}{2}} ) \ge 0 \\
  \Rightarrow tr(\tilde M^{\frac{p}{2}} ) - \frac{p}{2}tr(\tilde MM^{\frac{{p - 2}}{2}} ) \le \frac{{2 - p}}{2}tr(M^{\frac{p}{2}} ) \\
  \Rightarrow tr(\tilde M^{\frac{p}{2}} ) - \frac{p}{2}tr(\tilde MM^{\frac{{p - 2}}{2}} ) \le tr(M^{\frac{p}{2}} ) - \frac{p}{2}tr(MM^{\frac{{p - 2}}{2}} ), \\
 \end{array}
\]
which completes the proof. \hfill $\Box$
\begin{lemma}
\label{lemma4}
For any matrices $\tilde A,A$ with the same size and $\delta>0$, the following inequality holds when $0<p \le 2$.
\begin{equation}
\label{ineq4}
\begin{array}{l}
 tr((\tilde A^T \tilde A + \delta I)^{\frac{p}{2}} ) - \frac{p}{2}tr(\tilde A^T \tilde A(A^T A + \delta I)^{\frac{{p - 2}}{2}} ) \\
  \le tr((A^T A + \delta I)^{\frac{p}{2}} ) - \frac{p}{2}tr(A^T A(A^T A + \delta I)^{\frac{{p - 2}}{2}} ) \\
 \end{array}\,.
\end{equation}
\end{lemma}
\textbf{Proof.}
Note that $\tilde A^T \tilde A + \delta I$ and $A^T A + \delta I$ are positive definite matrices since $\delta>0$. Then according to Lemma \ref{lemma3} we have
\begin{equation}
  \begin{array}{l}
 tr((\tilde A^T \tilde A + \delta I)^{\frac{p}{2}} ) - \frac{p}{2}tr((\tilde A^T \tilde A + \delta I)(A^T A + \delta I)^{\frac{{p - 2}}{2}} ) \\
  \le tr((A^T A + \delta I)^{\frac{p}{2}} ) - \frac{p}{2}tr((A^T A + \delta I)(A^T A + \delta I)^{\frac{{p - 2}}{2}} ), \\
 \end{array}
\end{equation}
which indicates Eq.(\ref{ineq4}).
\hfill $\Box$

As a result, we have the following theorem.
\begin{theorem}
\label{thm1}
The Algorithm~\ref{alg1} will monotonically decrease the objective of the problem~(\ref{opgp1}) in each iteration until the algorithm converges.
\end{theorem}
\textbf{Proof}: In step 2 of Algorithm~\ref{alg1}, suppose the updated $x$ is $\tilde x$. According to step 2, we know
\begin{equation}
\label{ineq1}
f(\tilde x) + \mu\sum\limits_i {tr(g_i^T (\tilde x)g_i (\tilde x)D_i )}  \le f(x) + \mu\sum\limits_i {tr(g_i^T (x)g_i (x)D_i )},
\end{equation}
where the equality holds when and only when the algorithm converges.

For each $i$, according to Lemma~\ref{lemma4}, we have
\begin{small}
\begin{equation}\label{ineq20}
\begin{array}{l}
 tr((g_i^T (\tilde x)g_i (\tilde x) + \delta I)^{\frac{p}{2}} ) - \frac{p}{2}tr(g_i^T (\tilde x)g_i (\tilde x)(g_i^T (x)g_i (x) + \delta I)^{\frac{{p - 2}}{2}} ) \\
  \le tr((g_i^T (x)g_i (x) + \delta I)^{\frac{p}{2}} ) - \frac{p}{2}tr(g_i^T (x)g_i (x)(g_i^T (x)g_i (x) + \delta I)^{\frac{{p - 2}}{2}} ) \,.\\
\end{array}\,.
\end{equation}
\end{small}
Note that $D_i  = \frac{p}{2}(g_i^T (x)g_i (x) + \delta I)^{\frac{{p - 2}}{2}}$, so for each $i$ we have
\begin{equation}
\label{ineq21}
\begin{array}{l}
 tr((g_i^T (\tilde x)g_i (\tilde x) + \delta I)^{\frac{p}{2}} ) - tr(g_i^T (\tilde x)g_i (\tilde x)D_i ) \\
  \le tr((g_i^T (x)g_i (x) + \delta I)^{\frac{p}{2}} ) - tr(g_i^T (x)g_i (x)D_i ) \,.\\
 \end{array}
\end{equation}
Then we have
\begin{equation}
\label{ineq2}
\begin{array}{l}
 \sum\limits_i {tr((g_i^T (\tilde x)g_i (\tilde x) + \delta I)^{\frac{p}{2}} )}  - \sum\limits_i {tr(g_i^T (\tilde x)g_i (\tilde x)D_i )}  \\
  \le \sum\limits_i {tr((g_i^T (x)g_i (x) + \delta I)^{\frac{p}{2}} )}  - \sum\limits_i {tr(g_i^T (x)g_i (x)D_i )}  \\
 \end{array}\,.
\end{equation}
Summing Eq.~(\ref{ineq1}) and Eq.~(\ref{ineq2}) in the two sides, we arrive at
\begin{equation}\label{ineq3}
\begin{split}
     & f(\tilde x) + \mu\sum\limits_i {tr((g_i^T (\tilde x)g_i (\tilde x) + \delta I)^{\frac{p}{2}} )}   \\
     &\le f(x) + \mu\sum\limits_i {tr((g_i^T (x)g_i (x) + \delta I)^{\frac{p}{2}} )}\,.
\end{split}
\end{equation}
Note that the equality in Eq.(\ref{ineq3}) holds only when the algorithm converges. Thus the Algorithm~\ref{alg1} will monotonically decrease the objective of the problem (\ref{opgp1}) in each iteration until the algorithm converges. \hfill $\Box$

In the convergence, the equality in Eq.~(\ref{kkt}) will hold, thus the KKT condition \cite{boydConvex} of problem~(\ref{opgp1}) is satisfied.
Therefore, the Algorithm~\ref{alg1} will usually converge to a stationary point to the problem~(\ref{opgp1}). If the problem~(\ref{opgp1}) is convex,
the Algorithm~\ref{alg1} will usually converge to a global optimum solution.

\subsection{An Example Problem}

In this subsection, we give a concrete example and show how to derive the optimization algorithm for the example problem based on Algorithm~\ref{alg1}. The
problem is:
\begin{equation}
\label{opexamp}
  \mathop {\min }\limits_X \left\| {AX - Y} \right\|_{p,p}^p  + \mu_1\left\| {BX - Z} \right\|_{2,p}^p  + \mu_2\left\| X \right\|_{S_p }^p\,.
\end{equation}
This problem is a special case of problem~(\ref{opgp1}). According to Algorithm~\ref{alg1}, we only need to solve the following problem for
each $i$ in each iteration:
\begin{small}
\begin{equation}\label{opexampeasy}
\begin{split}
     & \mathop {\min }\limits_X \sum\limits_i {(Ax_i  - y_i )^T D_1^i (Ax_i  - y_i )} \\
     &   + \mu_1tr((BX - Z)^T D_2 (BX - Z)) + \mu_2 tr(X^T D_3 X),
\end{split}
\end{equation}
\end{small}
where $D_1^i$ is a diagonal matrix, the $k$-th diagonal element is $a^k x_i  - y_{ki}$,
$D_2$ is a diagonal matrix, the $k$-th diagonal element is $\frac{p}{2}((b^k X - z^k )^T (b^k X - z^k ) + \delta )^{\frac{{p - 2}}{2}}$,
$D_3  = \frac{p}{2}(XX^T  + \delta I)^{\frac{{p - 2}}{2}}$.

Taking the derivative of problem (\ref{opexampeasy}) w.r.t. $x_i$ and setting it to zero, we have
\begin{small}
\begin{equation}
\begin{array}{l}
 A^T D_1^i (Ax_i  - y_i ) + \mu_1B^T D_2 (Bx_i  - z_i ) + \mu_2 D_3 x_i  = 0 \\
  \Rightarrow \\
  x_i  = (A^T D_1^i A + \mu_1 B^T D_2 B + \mu_2 D_3 )^{ - 1} (A^T D_1^i y_i  + B^T D_2 z_i ) \,. \\
 \end{array}
\end{equation}
\end{small}
The detailed algorithm for solving the problem~(\ref{opexamp}) is listed in Algorithm~\ref{alg10}.

\begin{algorithm}
\label{alg10}
Initialize $x \in \mathcal{C}$, set $\delta$ be a very small constant  \;
\While{not converge}
{
1. For each $i$, calculate the diagonal matrix $D_1^i$, where  the $k$-th diagonal element is $\frac{p}{2}((a^k x_i  - y_{ki} )^2  + \delta )^{\frac{{p - 2}}{2}}$;
Calculate the diagonal matrix $D_2$, where the $k$-th diagonal element is $\frac{p}{2}((b^k X - z^k )^T (b^k X - z^k ) + \delta )^{\frac{{p - 2}}{2}}$;
Calculate the matrix $D_3  = \frac{p}{2}(XX^T  + \delta I)^{\frac{{p - 2}}{2}}$ \;
2. For each $i$, update $x_i$ by $x_i  = (A^T D_1^i A + \mu_1B^T D_2 B + \mu_2 D_3 )^{ - 1} (A^T D_1^i y_i  + B^T D_2 z_i )$
}
\KwOut{$x$.}
\caption{The algorithm to solve the problem~(\ref{opexamp}).}
\end{algorithm}

\section{Iteratively Reweighted Method for A More General Problem}

In this section, we focus on generalizing the problem~\eqref{opgp} to a more general problem as follows:
\begin{equation}
\label{opgp2}
\mathop {\min }\limits_{x \in \mathcal{C}} f(x) + \mu\sum\limits_i {h_i (g_i^T (x)g_i (x) + \delta I)},
\end{equation}
where $h_i(x)$ is an arbitrary \textbf{concave} and differentiable function. Inspired by the Algorithm~\ref{alg1}, the algorithm to solve the problem (\ref{opgp2}) is shown in Algorithm~\ref{alg2}, where we denote $\frac{{\partial h_i (g_i^T (x)g_i (x) + \delta I)}}{{\partial (g_i^T (x)g_i (x) + \delta I)}} = h'_i (g_i^T (x)g_i (x) + \delta I)$.
We will analyze the convergence of the algorithm in the next subsection.

For example, it can be easily checked that $h(M) = (tr(M))^\frac{p}{2}$, $h(M) = (tr(M))^\frac{p}{2}$ is concave and differentiable when $0<p \le 2$. So the Algorithm~\ref{alg2}
can be applied to solve the following problem:
\[
\mathop {\min }\limits_{x \in C} f(x) + \mu \sum\limits_i {(tr(g_i^T (x)g_i (x) + \delta I))^{\frac{p}{2}} }\,.
\]

\begin{algorithm}
\label{alg2}
Initialize $x \in \mathcal{C}$  \;
\While{not converge}
{ 1. For each $i$, calculate $D_i  = h'_i (g_i^T (x)g_i (x) + \delta I)$  \;
2. Update $x$ by solving the problem $\mathop {\min }\limits_{x \in \mathcal{C}} f(x) + \mu\sum\limits_i {tr(g_i^T (x)g_i (x)D_i )}$\;
}
\KwOut{$x$.}
\caption{The algorithm to solve the problem~(\ref{opgp2}).}
\end{algorithm}

\subsection{Convergence Analysis of Algorithm \ref{alg2}}

\begin{lemma}
\label{lemma5}
For an arbitrary concave and differentiable function $h(x)$, the following inequality holds:
\begin{equation}
\label{ineq5}
h(\tilde x) - h(x) \le h'(x)(\tilde x - x)\,.
\end{equation}
\end{lemma}
Then we have the following theorem.
\begin{theorem}
\label{thm2}
The Algorithm~\ref{alg1} will monotonically decrease the objective of the problem~(\ref{opgp2}) in each iteration until the algorithm converges.
\end{theorem}
\textbf{Proof}: In step 2 of Algorithm~\ref{alg2}, suppose the updated $x$ is $\tilde x$. According to step 2, we know
\begin{equation}
\label{ineq6}
f(\tilde x) + \mu\sum\limits_i {tr(g_i^T (\tilde x)g_i (\tilde x)D_i )}  \le f(x) + \mu\sum\limits_i {tr(g_i^T (x)g_i (x)D_i )},
\end{equation}
where the equality holds when and only when the algorithm converges.

Since $h_i(x)$ is concave for each $i$, according to Lemma~\ref{lemma5}, we have
\begin{equation}
\label{ineq70}
\begin{array}{l}
 h_i (g_i^T (\tilde x)g_i (\tilde x) + \delta I) - h_i (g_i^T (x)g_i (x) + \delta I) \\
  \le tr((g_i^T (\tilde x)g_i (\tilde x) - g_i^T (x)g_i (x))^T h'_i (g_i^T (x)g_i (x) + \delta I)) \\
 \end{array}\,.
\end{equation}
Note that $D_i  = h'_i (g_i^T (x)g_i (x) + \delta I)$, so for each $i$ we have
\begin{equation}
\begin{array}{l}
 h_i (g_i^T (\tilde x)g_i (\tilde x) + \delta I) - h_i (g_i^T (x)g_i (x) + \delta I) \\
  \le tr((g_i^T (\tilde x)g_i (\tilde x) - g_i^T (x)g_i (x))^T D_i ) \\
 \end{array}\,,
\end{equation}
and then
\begin{equation}
\begin{array}{l}
 h_i (g_i^T (\tilde x)g_i (\tilde x) + \delta I) - tr(g_i^T (\tilde x)g_i (\tilde x)D_i ) \\
  \le h_i (g_i^T (x)g_i (x) + \delta I) - tr(g_i^T (x)g_i (x)D_i ) \\
 \end{array}\,.
\end{equation}
So we have
\begin{equation}
\label{ineq7}
\begin{array}{l}
 \sum\limits_i {h_i (g_i^T (\tilde x)g_i (\tilde x) + \delta I)}  - \sum\limits_i {tr(g_i^T (\tilde x)g_i (\tilde x)D_i )}  \\
  \le \sum\limits_i {h_i (g_i^T (x)g_i (x) + \delta I)}  - \sum\limits_i {tr(g_i^T (x)g_i (x)D_i )}  \\
 \end{array}\,.
\end{equation}
Summing Eq.~(\ref{ineq6}) and Eq.~(\ref{ineq7}) in the two sides, we arrive at
\begin{equation}\label{ineq8}
\begin{split}
     & f(\tilde x) + \mu\sum\limits_i {h_i (g_i^T (\tilde x)g_i (\tilde x) + \delta I)} \\
     &   \le f(x) + \mu \sum\limits_i {h_i (g_i^T (x)g_i (x) + \delta I)}
\end{split}
\end{equation}
Note that the equality in Eq.(\ref{ineq3}) holds only when the algorithm converges. Thus the Algorithm~\ref{alg1} will monotonically decrease the objective of the problem (\ref{opgp1}) in each iteration until the algorithm converges. \hfill $\Box$

\begin{lemma}
\label{lem6}
Suppose $g (x)$ is a scalar, vector or matrix output function, $x$ is a scalar, vector or matrix variable, then we have
\begin{equation}
\frac{{\partial h(g^T (x)g(x) + \delta I)}}{{\partial x}} = \frac{{tr\left( {2h'(g^T (x)g(x) + \delta I)g^T (x)\partial g(x)} \right)}}{{\partial x}}\,.
\end{equation}
\end{lemma}
\textbf{Proof}:
According to the chain rule in Lemma \ref{lem1}, we have
\[
\begin{array}{l}
 \frac{{\partial h(g^T (x)g(x) + \delta I)}}{{\partial x}} = \frac{{tr\left( {\left( {h'(g^T (x)g(x) + \delta I)} \right)^T \partial (g^T (x)g(x) + \delta I)} \right)}}{{\partial x}} \\
  = \frac{{tr\left( {h'(g^T (x)g(x) + \delta I)^T (\partial (g^T (x))g(x) + g^T (x)\partial g(x))} \right)}}{{\partial x}} \\
  = \frac{{tr\left( {h'(g^T (x)g(x) + \delta I)(\partial g(x))^T g(x) + h'(g^T (x)g(x) + \delta I)g^T (x)\partial g(x))} \right)}}{{\partial x}} \\
  = \frac{{tr\left( {h'(g^T (x)g(x) + \delta I)g^T (x)(\partial g(x)) + h'(g^T (x)g(x) + \delta I)g^T (x)\partial g(x))} \right)}}{{\partial x}} \\
  = \frac{{tr\left( {2h'(g^T (x)g(x) + \delta I)g^T (x)\partial g(x)} \right)}}{{\partial x}} \\
 \end{array}
\]
which completes the proof. \hfill $\Box$

\begin{theorem}
\label{thm3}
The Algorithm~\ref{alg2} will converge to the KKT condition of the problem~(\ref{opgp2}).
\end{theorem}
\textbf{Proof}:
The Lagrangian function of the problem (\ref{opgp2}) is
\begin{equation}
\mathcal{L}_1 (x,\lambda ) = f(x) + \mu\sum\limits_i {h_i (g_i^T (x)g_i (x) + \delta I)}  - r(x,\lambda )
\end{equation}
Based on the KKT condition, by setting the derivative of $L_1 (x,\lambda )$ w.r.t. $x$, we have
\begin{equation}\label{kkteq00}
\frac{{\partial \mathcal{L}_1 (x,\lambda )}}{{\partial x}} = f'(x) + \mu\sum\limits_i {\frac{{\partial h_i (g_i^T (x)g_i (x) + \delta I)}}{{\partial x}}}  - \frac{{\partial r(x,\lambda )}}{{\partial x}} = 0
\end{equation}
According to Lemma (\ref{lem6}), Eq.(\ref{kkteq00}) can be rewritten as
\begin{equation}\label{kkteq0}
\begin{split}
     & \frac{{\partial \mathcal{L}_1 (x,\lambda )}}{{\partial x}} = f'(x)+\\
     &    \mu \sum\limits_i {\frac{{tr\left( {2h'_i (g_i^T (x)g_i (x) + \delta I)g_i^T (x)\partial g_i (x)} \right)}}{{\partial x}}}  - \frac{{\partial r(x,\lambda )}}{{\partial x}} = 0\,.
\end{split}
\end{equation}

On the other hand, in the second step of the Algorithm~\ref{alg2}, we solve the problem $\mathop {\min }\limits_{x \in \mathcal{C}} f(x) + \sum\limits_i {tr(g_i^T (x)g_i (x)D_i )}$. The Lagrangian function of this problem is
\begin{equation}
\mathcal{L}_2 (x,\lambda ) = f(x) + \mu \sum\limits_i {tr(g_i^T (x)g_i (x)D_i )}  - r(x,\lambda )\,.
\end{equation}
By setting the derivative of $\mathcal{L}_2 (x,\lambda )$ w.r.t. $x$, we have
\begin{equation}
\label{kkteq}
\frac{{\partial \mathcal{L}_2 (x,\lambda )}}{{\partial x}} = f'(x) + \mu\sum\limits_i {\frac{{\partial tr(g_i^T (x)g_i (x)D_i )}}{{\partial x}}}  - \frac{{\partial r(x,\lambda )}}{{\partial x}} = 0\,.
\end{equation}
According to Lemma (\ref{lem3}), Eq.(\ref{kkteq}) can be rewritten as
\begin{equation}
\label{kkteq1}
f'(x) + \mu\sum\limits_i {\frac{{tr\left( {2D_i g_i^T (x)\partial g_i (x))} \right)}}{{\partial x}}}  - \frac{{\partial r(x,\lambda )}}{{\partial x}} = 0\,.
\end{equation}
Thus we find a solution satisfying Eq.(\ref{kkteq1}) in each iteration according to the second step of Algorithm~\ref{alg2}. In the convergence of the Algorithm~\ref{alg2}, note that $D_i  = h'_i (g_i^T (x)g_i (x) + \delta I)$ according to the first step of Algorithm~\ref{alg2}, Eq.(\ref{kkteq1}) is
equivalent to
\begin{equation}\label{kkteq2}
\begin{split}
     & f'(x) + \mu\sum\limits_i {\frac{{tr\left( {2h'_i (g_i^T (x)g_i (x) + \delta I)g_i^T (x)\partial g_i (x))} \right)}}{{\partial x}}} \\
     & - \frac{{\partial r(x,\lambda )}}{{\partial x}} = 0 \,.
\end{split}
\end{equation}
Therefore, the solution $x$ satisfies Eq.(\ref{kkteq2}) in the convergence of the Algorithm~\ref{alg2}, which is exactly the same as the KKT condition of the problem~(\ref{opgp2}) in Eq.(\ref{kkteq0}).
\hfill $\Box$

Theorem \ref{thm2} and \ref{thm3} indicate that the Algorithm~\ref{alg2} will converge, and usually converge to a stationary point to the problem~(\ref{opgp2}). If the problem~(\ref{opgp2}) is convex, the Algorithm~\ref{alg1} will usually converge to a global optimum solution.

It is worth to pointing out that the similar algorithm and results can also be found for the following general problem:
\begin{equation}
\mathop {\min }\limits_{x \in \mathcal{C}} f(x) + \mu\sum\limits_i {h_i (g_i(x))},
\end{equation}
where $h_i(x)$ is an arbitrary \textbf{concave} and differentiable function. In this case, the two steps in Algorithm~\ref{alg2} becomes $D_i  = h'_i (g_i (x))$ and $\mathop {\min }\limits_{x \in \mathcal{C}} f(x) + \sum\limits_i {tr((g_i (x))^TD_i )}$, respectively.

\section{Experimental Results}

In this section, we will conduct diversified experiments to empirically demonstrate the convergence rate as well as the computing accuracy of our new algorithm.

\subsection{Data Description}

A total of $16$ publicly reachable real benchmark data sets have participated in our evaluations, including: AR10P, PIX10P, PIE10P [\cite{sim2002cmu}], ORL10P [\cite{samaria1994parameterisation}], ALLAML [\cite{fodor1997dna}], MLLML [\cite{armstrong2001mll}], LUNG [\cite{bhattacharjee2001classification}], Prostate-GE [\cite{singh2002gene}], Carcinomas [\cite{su2001molecular}, \cite{yang2006stable}], GLIOMA [\cite{nutt2003gene}], CLL-SUB-111 [\cite{haslinger2004microarray}], TOX-171 [\cite{kwissa2014dengue}], SMK-CAN-187 [\cite{spira2007airway}], Prostate-MS [\cite{petricoin2002serum}], ARCENE and DBWorld, among which the first four are face image data sets\footnote{Downloaded from http://featureselection.asu.edu/datasets.php}, the next eleven are gene expression data sets, while the last two are life data sets achieved from the UCI Repository[\cite{BacheLichman:2013}].
Detailed property of these 17 data sets is introduced as below.

\textbf{AR10P} data set records $130$ face images from $10$ different people, with each person contributing $13$ images to the data set. The faces are represented by $60*40$ pixel images, thus the dimensionality of each sample is $2400$. This data set has participated in numerous face recognition experiments.

\textbf{PIX10P} data set consists of $100$ face images from $5$ male and $5$ female people. For each participant, $10$ face images with the dimensionality of $100*100$ are included. This is also a famous data set utilized in face recognition simulation.

\textbf{PIE10P} data set is collected by the Robotics Institute of Carnegie Mellon University. It is composed of 210 face images from 10 different people, with 21 faces from each testee. Each face is depicted by a 55*44 image. Similar to the previous two data sets, this data set also shows up in face recognition experiments with high frequency.

\textbf{ORL10P} data set is also known as the ''AT\&T face data sets". It collects 400 face images from 40 distinct subjects. All images are of the size 92*112 pixels, with 256 grey levels per pixel. In our experiments, we use the selected data from the ASU Feature Selection Database where all 10 classes without glasses are included.

\textbf{ALLAML} data set records 7129 genes (sequences) information from the Affymetrix 6800 chip. It has a total of 72 samples in two classes, ALL and AML, of 47 and 25 samples, respectively.

\textbf{MLLML} data set is downloaded from Liubjana A.I. lab website, which contains a subset of human acute lymphoblastic leukemias with a chromosomal translocation involving the mixed-lineage leukemia gene. As is shown in the data set, the mixed-lineage leukemia (MLL) gene has a clear pattern to be separated from ALL and AML, thus this data set has been widely utilized in classification experiments. This data set is composed of 72 samples from three classes, which are ALL, AML and MLL. The number of samples of these three classes are 24, 28 and 20, respectively. Each sample has 12582 genes.


\textbf{LUNG} data set provides a source for the study of lung cancer. It has 203 samples in five classes, among which there are 139 adenocarcinoma (AD), 17 normal lung (NL), 6 small cell lung cancer (SMCL), 21 squamous cell carcinoma (SQ) as well as 20 pulmonary carcinoid (COID) samples. Each sample has 3312 genes.

\textbf{Prostate-GE} data set records gene information of both prostate cancer and normal patients. It contains 102 samples of two classes, among which there are 52 tumor samples and 50 normal samples, respectively. In our experiment, each sample has 5966 genes.

\textbf{Carcinomas} data set shows the influence of genes on various types of carcinomas. This data set contains 174 samples of 11 classes, which are 26 samples of prostate carcinoma, 8 samples of bladder/ureter carcinoma, 26 samples of breast carcinoma, 23 samples of colorectal carcinoma, 12 samples of gastroesophagus carcinoma, 11 samples of kidney carcinoma, 7 samples of liver carcinoma, 27 samples of ovary carcinoma, 6 samples of pancreas carcinoma, 14 samples of lung adeno-carcinoma and 14 samples of lung squamous cell carcinoma. Each sample contains 9182 genes as features.

\textbf{GLIOMA} data set encompasses 50 samples of four different disease statuses, where there are 14 cancer glioblastomas (CG), 14 noncancer glioblastomas (NG), 7 cancer oligodendrogliomas (CO) and 15 non-cancer oligodendrogliomas (NO) samples, respectively. Each sample is described by 4433 genes.

\textbf{CLL-SUB-111} data set composes of microarray gene expression information of 111 samples from 3 classes. It provides array analysis results on chronic lymphocytic leukemia (CLL) patients. Each sample contains 11340 genes.

\textbf{TOX-171} data set records blood analysis of acute Dengue virus (DENV) patients, which provides references for molecular mechanisms studies of DENV infection. This data set consists of 171 samples from 4 classes, where each sample have 5748 features.

\textbf{SMK-CAN-187} data set provides insights into the study of lung cancer inducement. It records RNA microarray information from 187 samples of two classes, including Bronchial Epithelium of Smokers with Lung Cancer and those without. Each sample has a total of 19993 features.

\textbf{Prostate-MS} data set contains a total of 332 samples from three different classes, which are 69 samples diagnosed as prostate cancer, 190 samples of benign prostate hyperplasia, as well as 63 normal samples showing no evidence of disease. Each sample has 15154 genes.

\textbf{ARCENE} data set provides mass-spectrometric information for both cancer and normal patterns. The size of this data set is 100, where each sample has a total of 10000 attributes. It provides challenge for two-class classification with continuous input data.

\textbf{DBWorld} contains 64 emails manually collected from DBWorld mailing list. These emails are classified in two classes: announces of conferences and everything else. Each email is depicted by 4702 features in bag-of-words representation.

\subsection{Experiments on Solving the Example Problem (\ref{opexamp}) }\label{sec:test1}

In this experiment, we examine the efficiency of our algorithm for solving Problem (\ref{opexamp}) with different values of $p$.

There are four data sets required as input in this experiment, that is, $A$ and $Y$, $B$ and $Z$, where $A$ and $B$, $Y$ and $Z$ are required to have the same dimensionality. However, it's tough to find real benchmark data sets with exactly the same dimensionality. But experiments on purely synthetic data lack challenges to some extent. Hence we decide to combine benchmark data sets with synthetic data. The data we utilized for matrix $A$ and $Y$ are real benchmark data sets, while for matrix $B$ and $Z$ are synthetic data obeying Gaussian distribution, whose dimensionality are set to be the same with the corresponding real benchmark data.

Since the purpose of this experiment is to show the convergence performance of our method with different $p$ values, here we choose seven disparate $p$ values in the range of $0 < p \leq 2$ which are \{0.1, 0.5, 0.8, 1, 1.2, 1.5, 2\}. We performed experiments on eight data sets with comparatively small dimensionality, which are AR10P, PIE10P, ALLAML, LUNG, Prostate-GE, Carcinomas, GLIOMA, and TOX-171.

\begin{figure*}[!t]
\label{fig1} 
  \centering
  \subfigure[dbworld]{
    \includegraphics[width=0.23\textwidth]{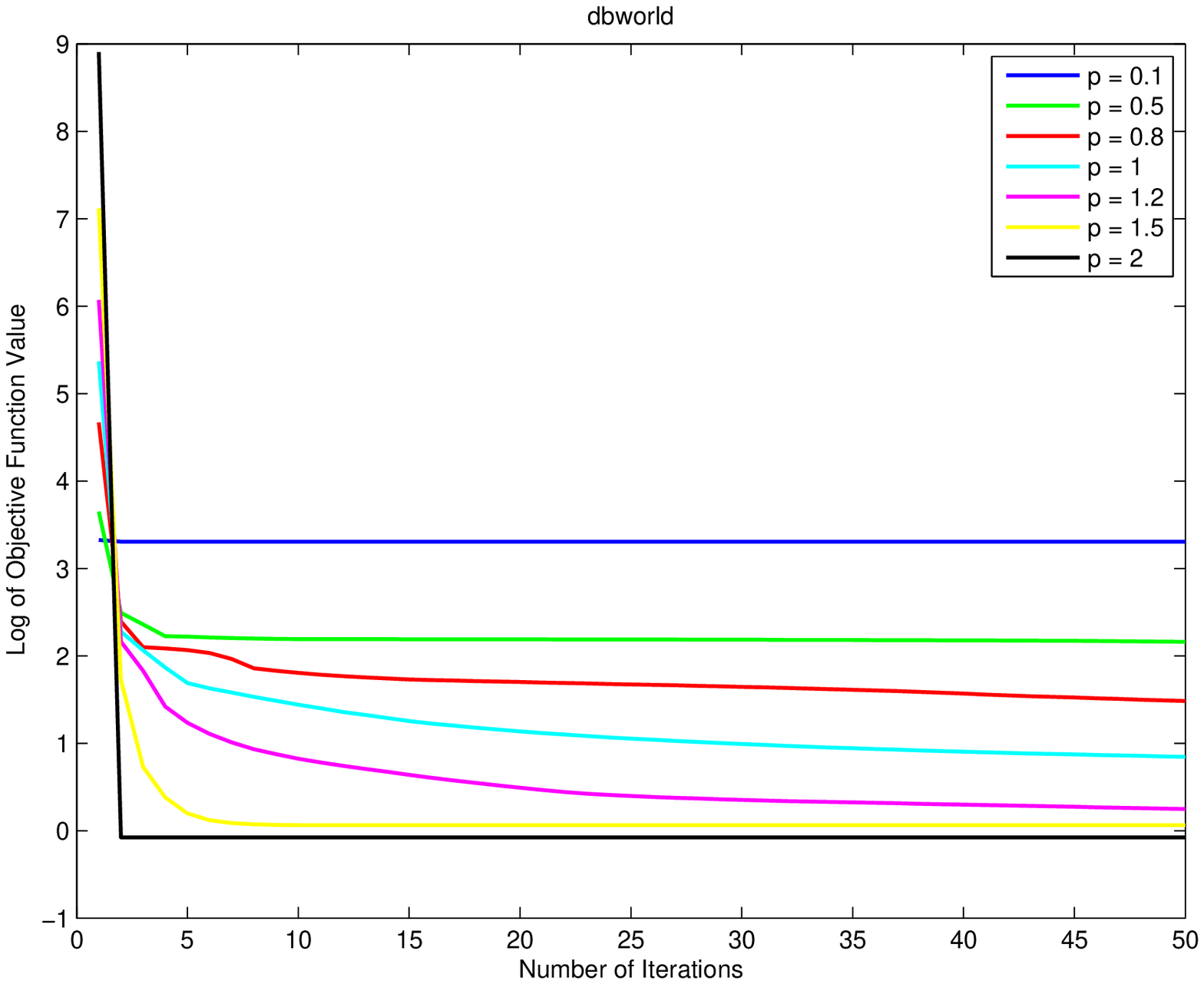}}
  \subfigure[Prostate−GE]{
    \includegraphics[width=0.23\textwidth]{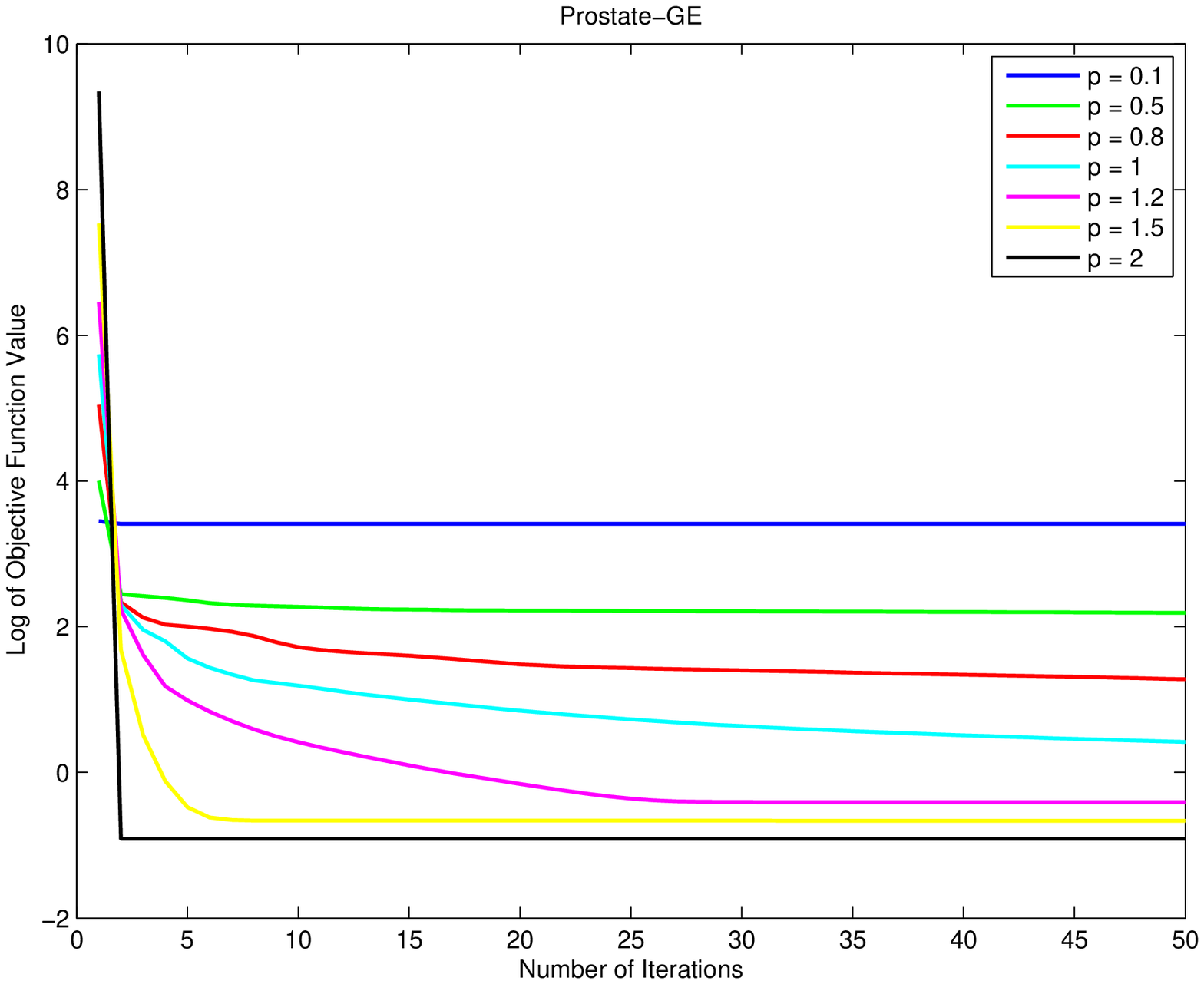}}
  \subfigure[Carcinomas]{
    \includegraphics[width=0.23\textwidth]{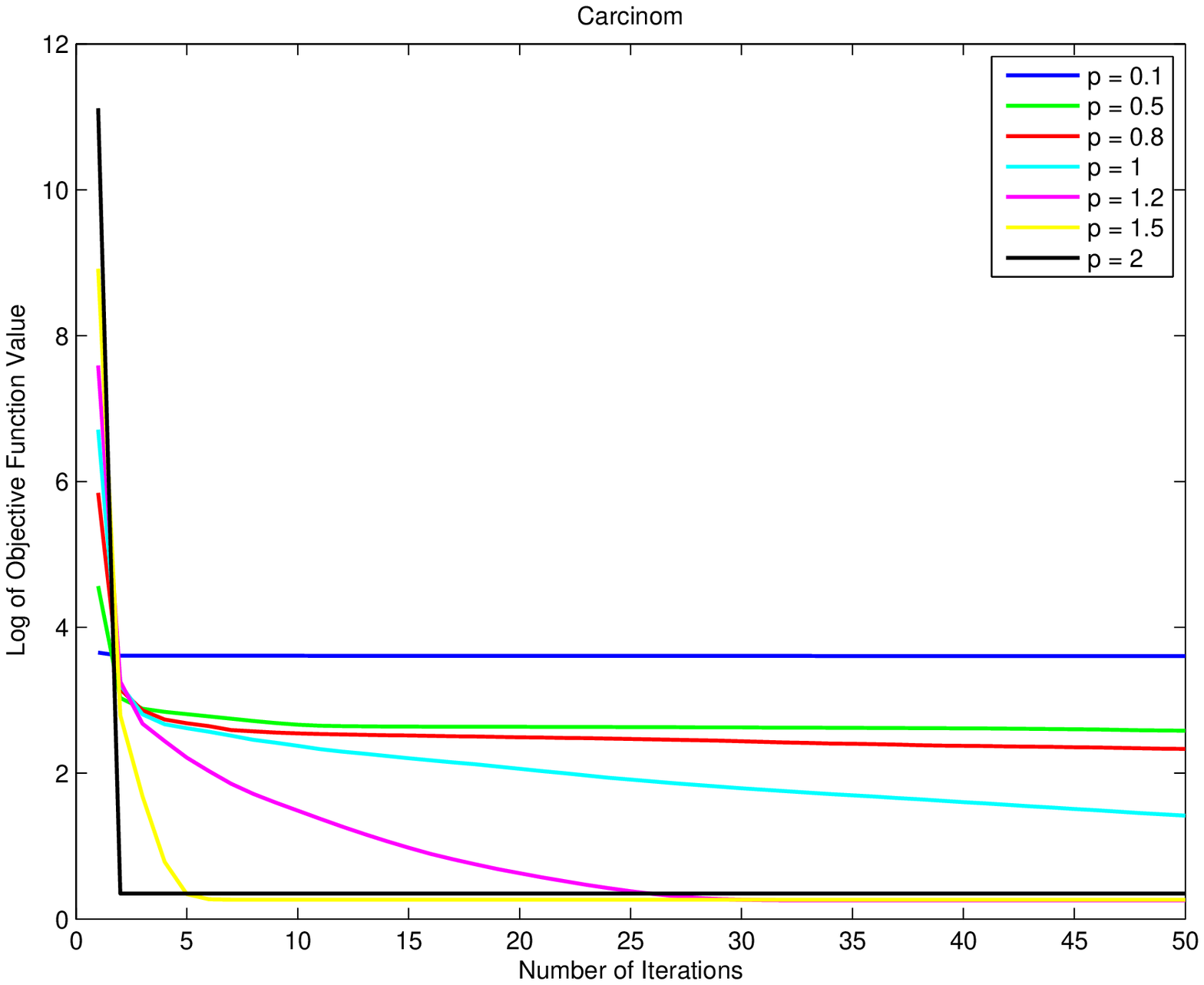}}
  \subfigure[GLIOMA]{
    \includegraphics[width=0.23\textwidth]{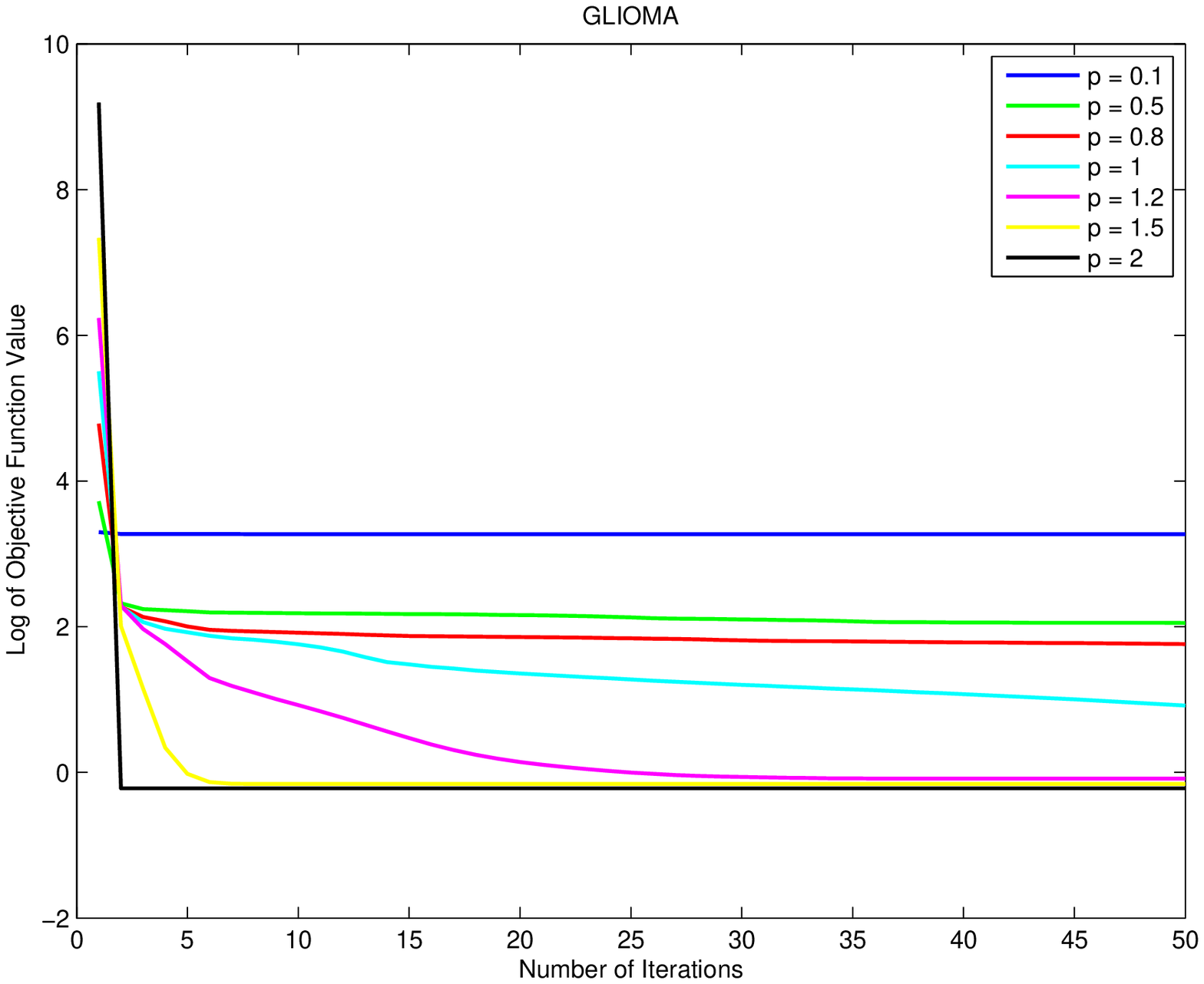}}
  \subfigure[LUNG]{
    \includegraphics[width=0.23\textwidth]{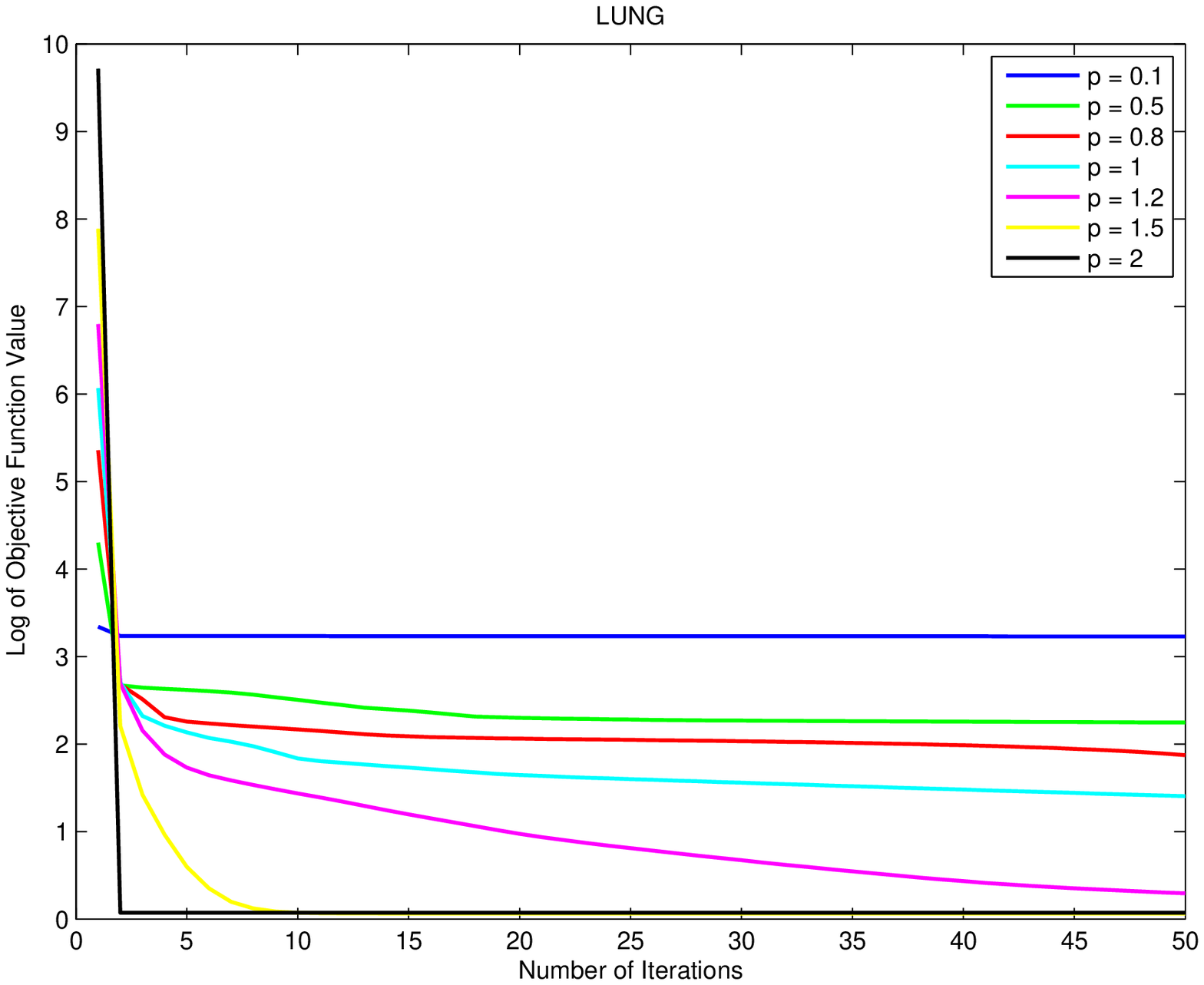}}
  \subfigure[ALLAML]{
    \includegraphics[width=0.23\textwidth]{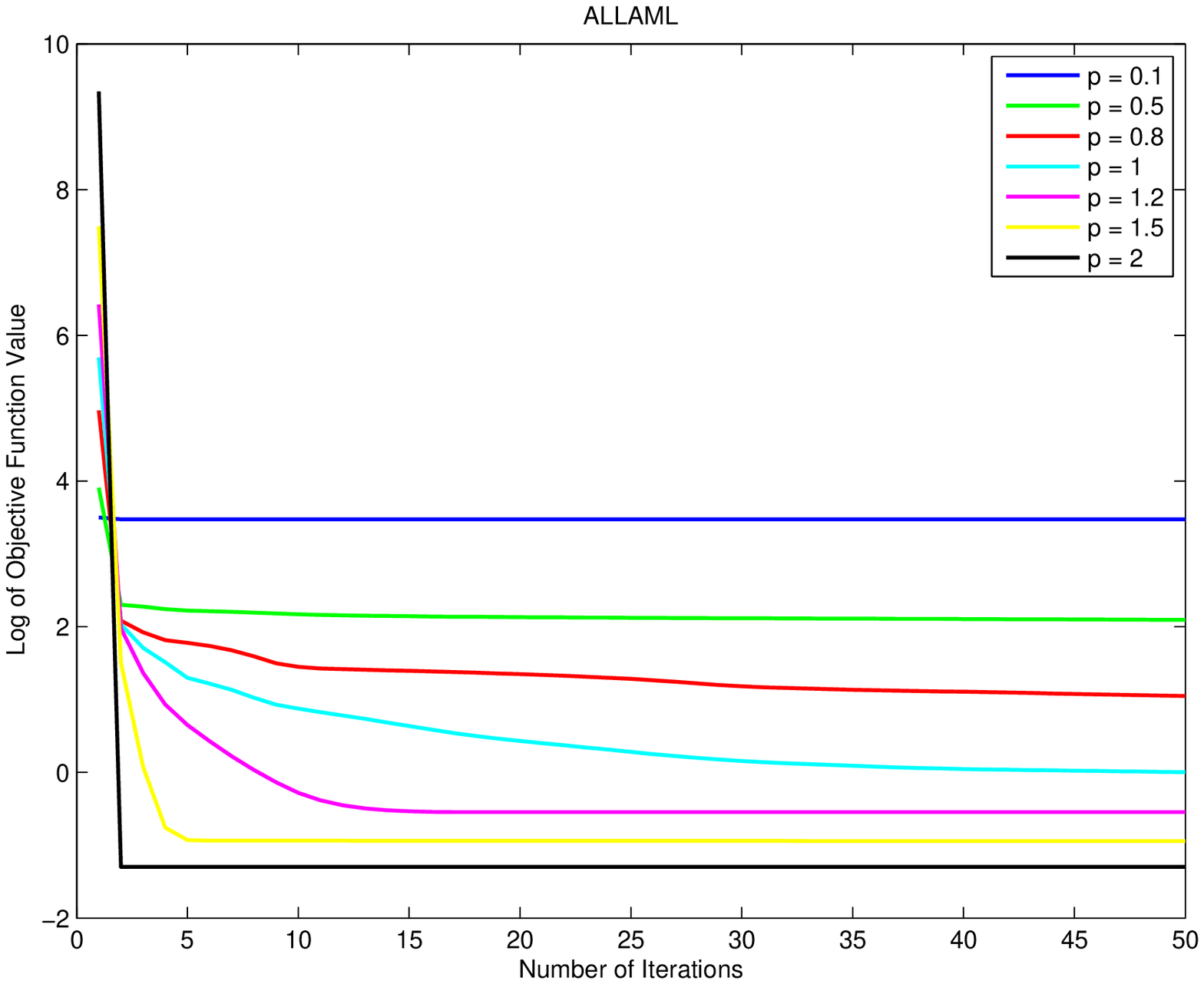}}
  \subfigure[TOX−171]{
    \includegraphics[width=0.23\textwidth]{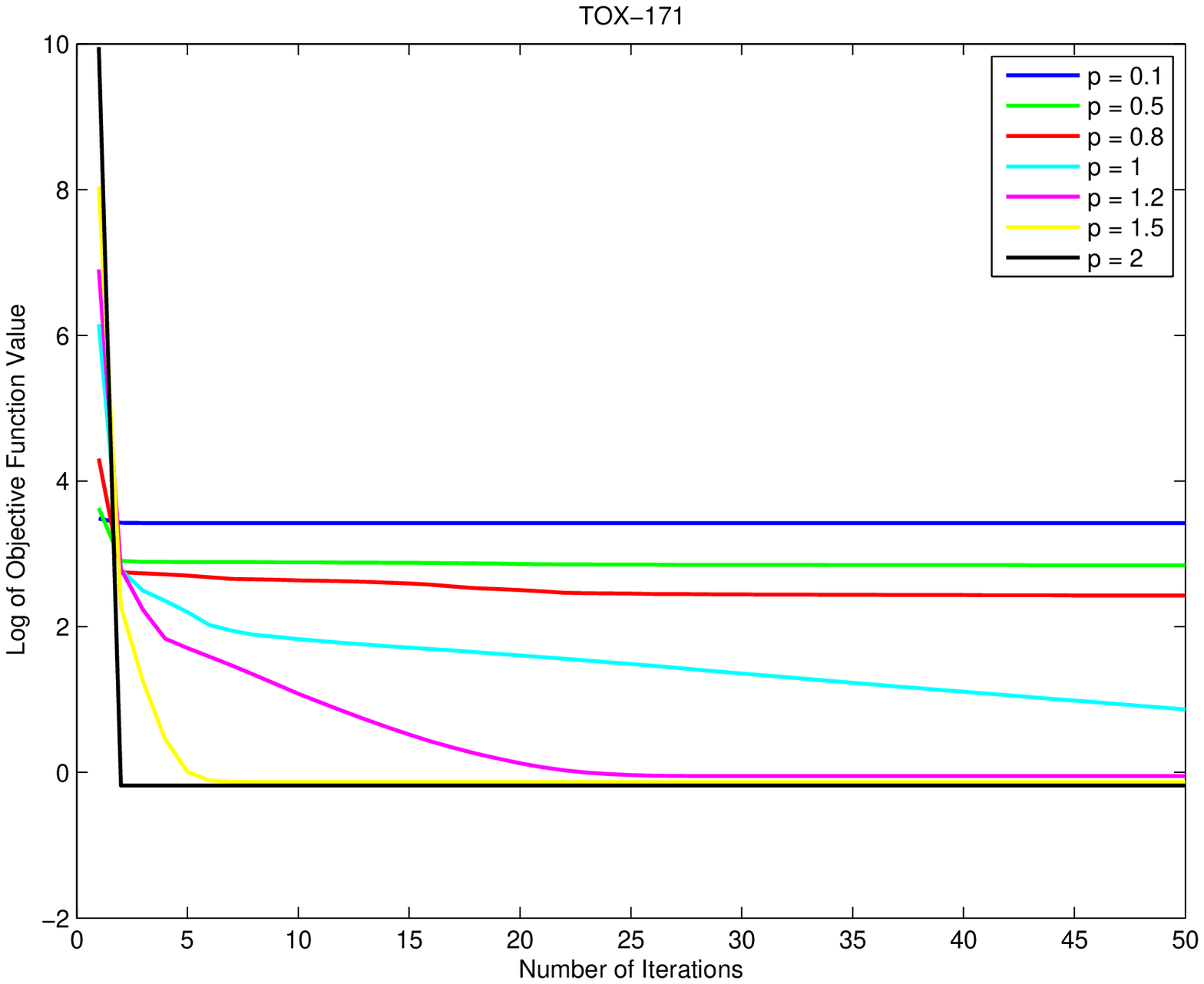}}
  \subfigure[warpAR10P]{
    \includegraphics[width=0.23\textwidth]{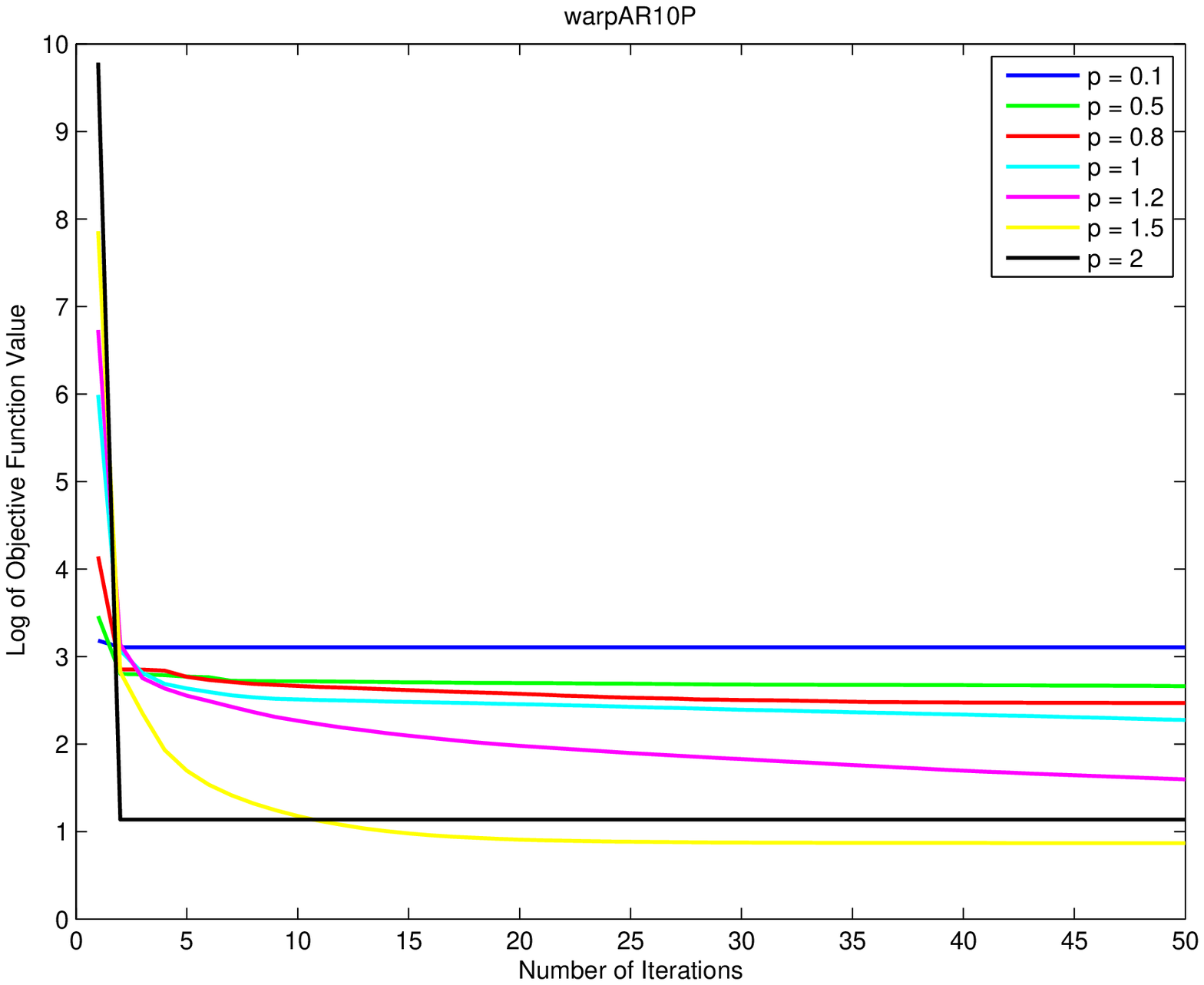}}
\vbox{}
\caption{Log of objective function value with different $p$ value. The objective function is Eq.~\eqref{opexamp}.}
\end{figure*}

The results are displayed in Fig.~\ref{fig1}, from which we know that our methods converge very fast, usually within $50$ iterations. Especially when $p$ = 2, our method converges in just one iteration. That's because when $p$ = 2, Problem (\ref{opexamp}) becomes:
\begin{equation}
\label{ex1}
  \mathop {\min }\limits_X \left\| {AX - Y} \right\|_{F}^2  + \mu_1\left\| {BX - Z} \right\|_{F}^2  + \mu_2tr(X^{T}X)\,.
\end{equation}

In Problem (\ref{ex1}), if we take derivative w.r.t. $X$ and set it to zero, we will get:
\begin{equation}
\label{ex2}
  (A^{T}A + \mu_1B^{T}B + \mu_2I)X = A^{T}Y + B^{T}Z\,.
\end{equation}
where $X$ has a closed form solution, thus our method converges in just one iteration.

\subsection{Experiments on the Proximal Problem}

This experimental subsection talks about solving another complex problem as below:
\begin{equation}
\label{ex2ori}
\mathop {\min }\limits_X f(X)  + \gamma _1 \left\| X \right\|_{p,p}^p  + \gamma _2 \left\| X \right\|_{2,p}^p  + \gamma _3 \left\| X \right\|_{S_p }^p
\end{equation}

According to the series of work by Yurii Nesterov \cite{nesterov1983method,nesterovintroductory,nesterov2005smooth,nesterov2007gradient}, we can solve Problem (\ref{ex2ori}) via the proximal method.
Before directly going to the solving process, let's first have a brief introduction on the proximal method.

For a general minimization problem w.r.t. $x$ as follows:
\begin{equation}
\label{pmori}
\mathop {\min }\limits_x f(x) + \varphi (x)
\end{equation}

We can obtain an approximate equality of function $f(x)$ according to its Taylor series:
\[f(x) \approx f({x_{t - 1}}) + tr({(x - {x_{t - 1}})^T}f'({x_{t - 1}})) + \frac{L}{2}\left\| {x - {x_{t - 1}}} \right\|_F^2\]
where $L = f''({x_{t - 1}})$.

Then the original equation in Problem (\ref{pmori}) can be rewritten as:
\begin{equation}
\begin{split}
&f(x) + \varphi (x) \approx ~f({x_{t - 1}}) +\\
&  tr({(x - {x_{t - 1}})^T}f'({x_{t - 1}})) + \frac{L}{2}\left\| {x - {x_{t - 1}}} \right\|_F^2 + \varphi (x)\\
 & = ~\frac{L}{2}\left\| {x - ({x_{t - 1}} - \frac{1}{L}f'({x_{t - 1}}))} \right\|_F^2 + \varphi (x)
\end{split}
\end{equation}
thus we can update $x_t$ in each iteration as the optimal solution to the following problem:
\begin{equation}
{x_t} = \arg \mathop {\min }\limits_x \frac{L}{2}\left\| {x - ({x_{t - 1}} - \frac{1}{L}f'({x_{t - 1}}))} \right\|_F^2 + \varphi (x)\,.
\end{equation}

It has been proven in Yurii Nesterov's work that if the original problem is convex, the proximal method will reach its global optimum with a convergence rate $O(\frac{1}{t})$; otherwise it will end up with a stationary point.

Based on the proximal method introduced above, we can optimize Problem (\ref{ex2ori}) by solving the following problem in each iteration:
\begin{equation}
\label{ex2}
\mathop {\min }\limits_X \left\| {X - V} \right\|_F^2  + \gamma'_1 \left\| X \right\|_{p,p}^p  + \gamma'_2 \left\| X \right\|_{2,p}^p  + \gamma'_3 \left\| X \right\|_{S_p }^p\,,
\end{equation}
where $V = {X_{t - 1}} - \frac{1}{L}f'({X_{t - 1}})$, $ \gamma'_1 = \frac{2\gamma_1}{L}$, $\gamma'_2 = \frac{2\gamma_2}{L}$ and $\gamma'_3 = \frac{2\gamma_3}{L}$.

It's apparent that Problem (\ref{ex2}) can solved using our new algorithm. So in this subsection our goal is to check the efficiency of our algorithm for solving problem (\ref{ex2}).

In this experiment we utilized all the eight data sets used in Sect.~\ref{sec:test1}, and varied $p$ in the set  $\{0.1, 0.5, 0.8, 1, 1.2, 1.5, 2\}$. As for the parameter $\gamma'_1$, $\gamma'_2$ and $\gamma'_3$, here we simply set them to be 1 as we are just devoted to testing the convergence rate in this experiment. If the purpose is instead to minimize Problem (\ref{ex2}) and find a suitable $X$ that best accomplishes a certain task, tuning $\gamma'_1$, $\gamma'_2$ and $\gamma'_3$ provides a convenient way for improving the performance.
We present the results on disparate data sets in Fig.~2.

Obviously, our methods converge very fast, almost all within $20$ iterations. And we also witness a special case where $p$ = 2, that our method converges in just one iteration. The reason is similar to above, i.e., when $p$ = 2, problem (\ref{ex2}) has a closed form solution, which urges our algorithm to converge in merely one iteration.

\begin{figure*}[!t]
\label{fig2} 
  \centering
  \subfigure[DBWorld]{
    \includegraphics[width=0.23\textwidth]{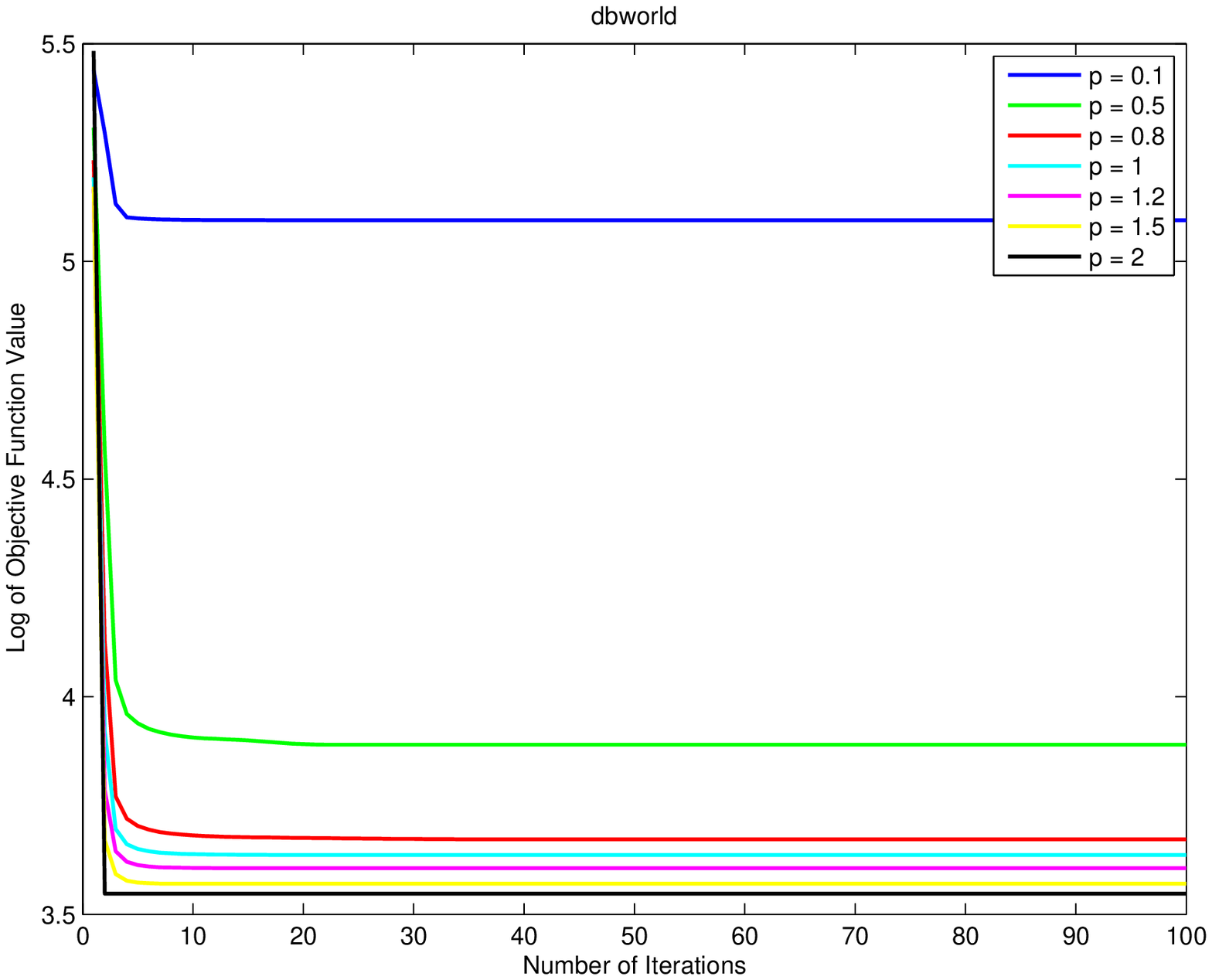}}
  \subfigure[Prostate−GE]{
    \includegraphics[width=0.23\textwidth]{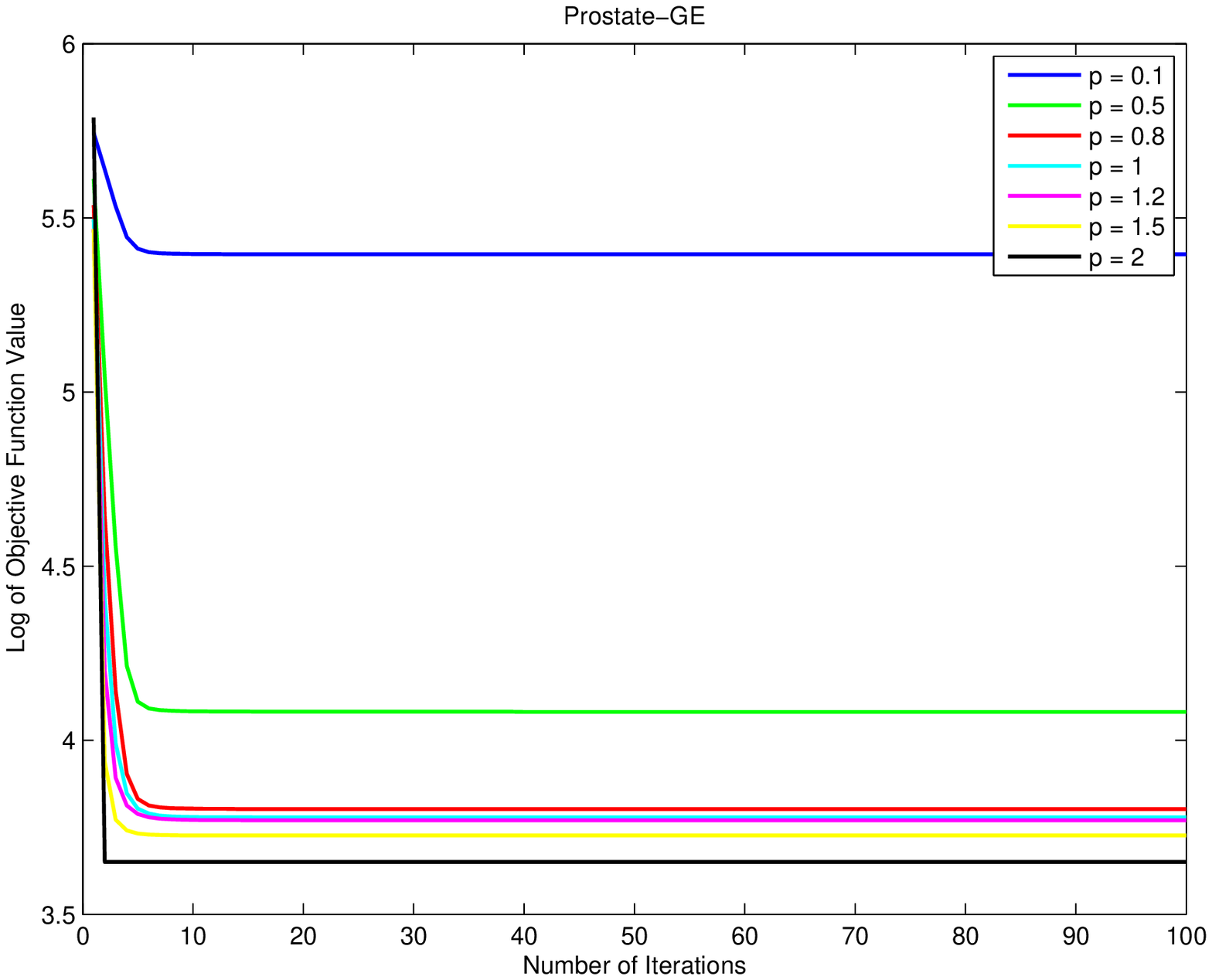}}
  \subfigure[Carcinomas]{
    \includegraphics[width=0.23\textwidth]{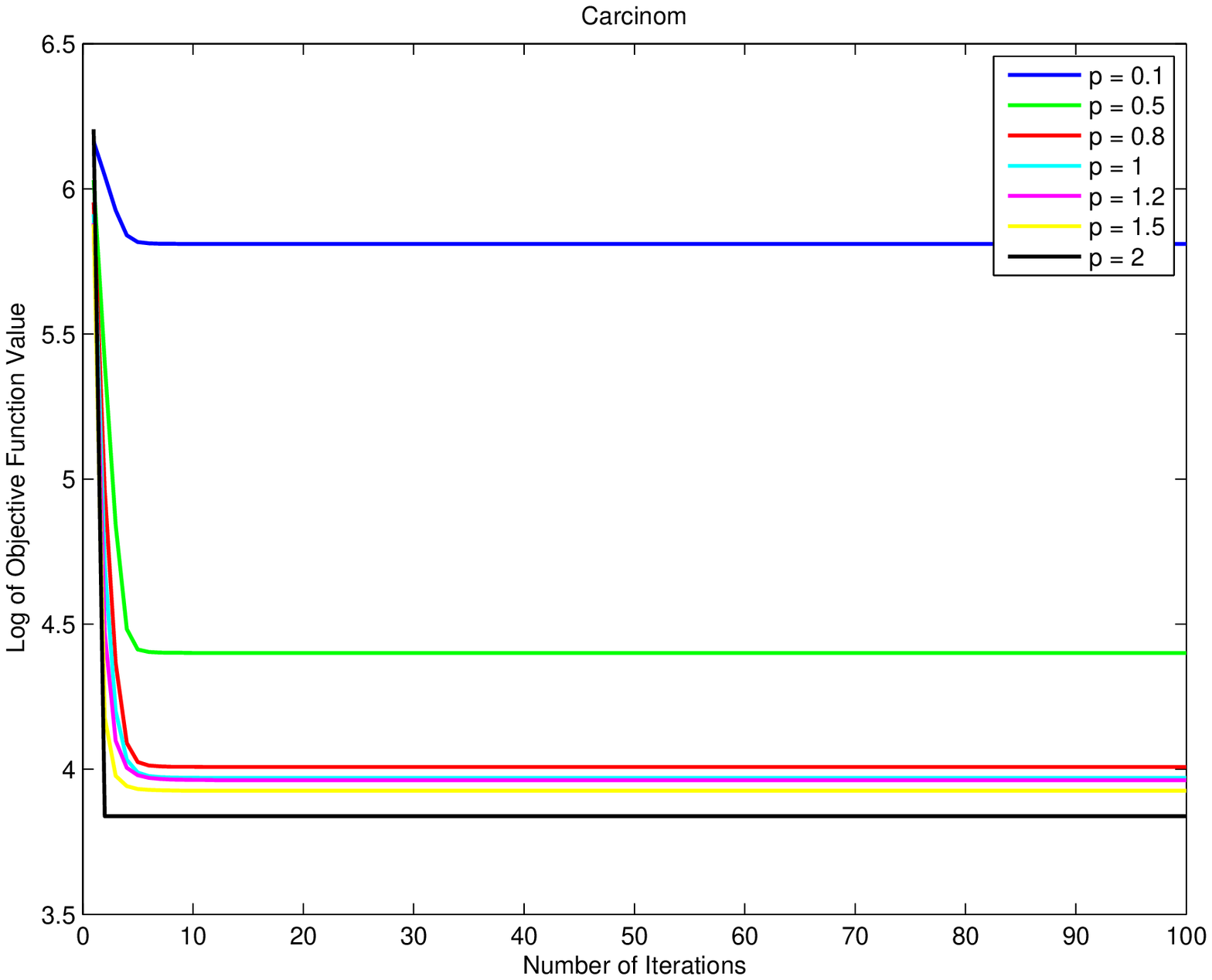}}
  \subfigure[GLIOMA]{
    \includegraphics[width=0.23\textwidth]{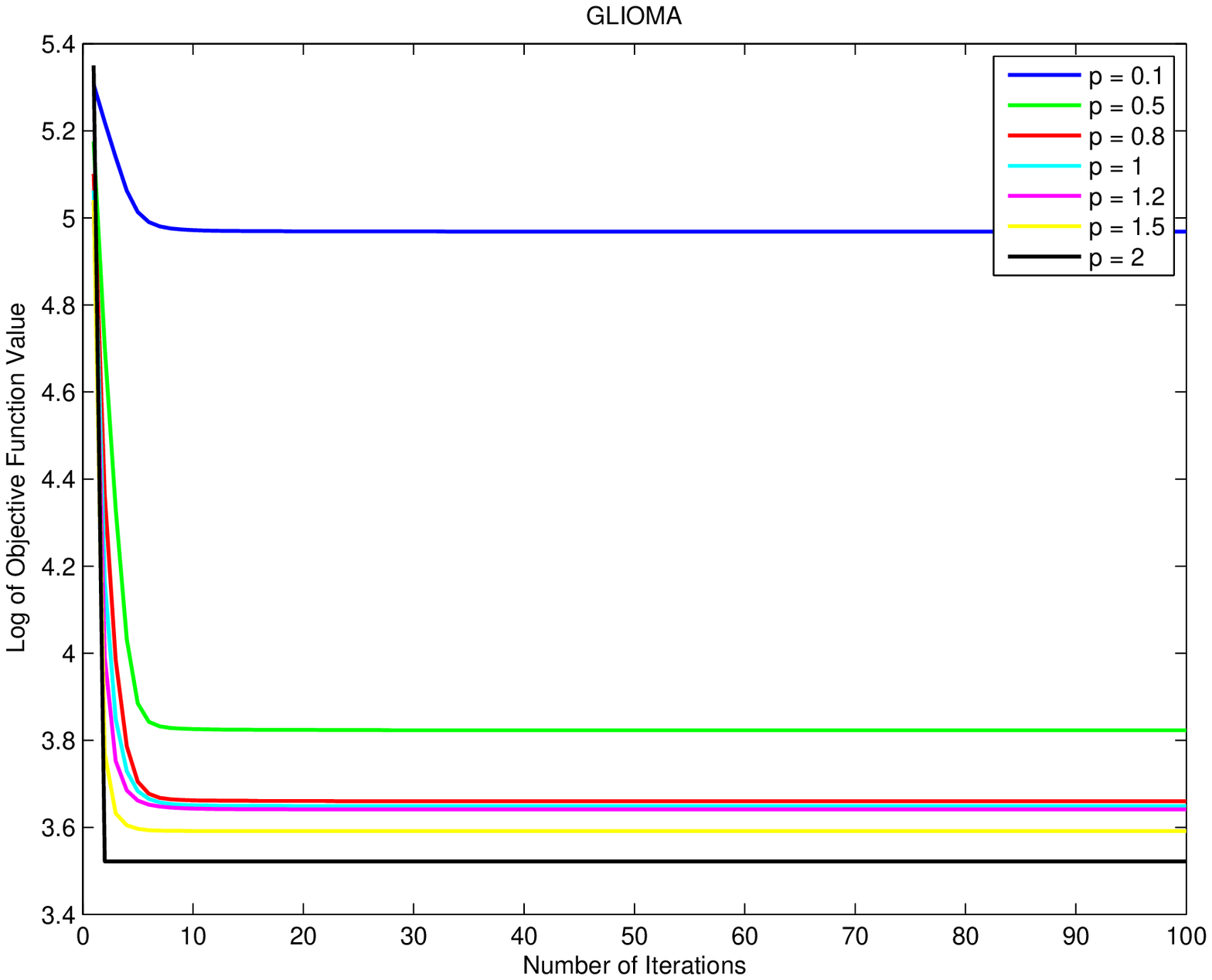}}
  \subfigure[LUNG]{
    \includegraphics[width=0.23\textwidth]{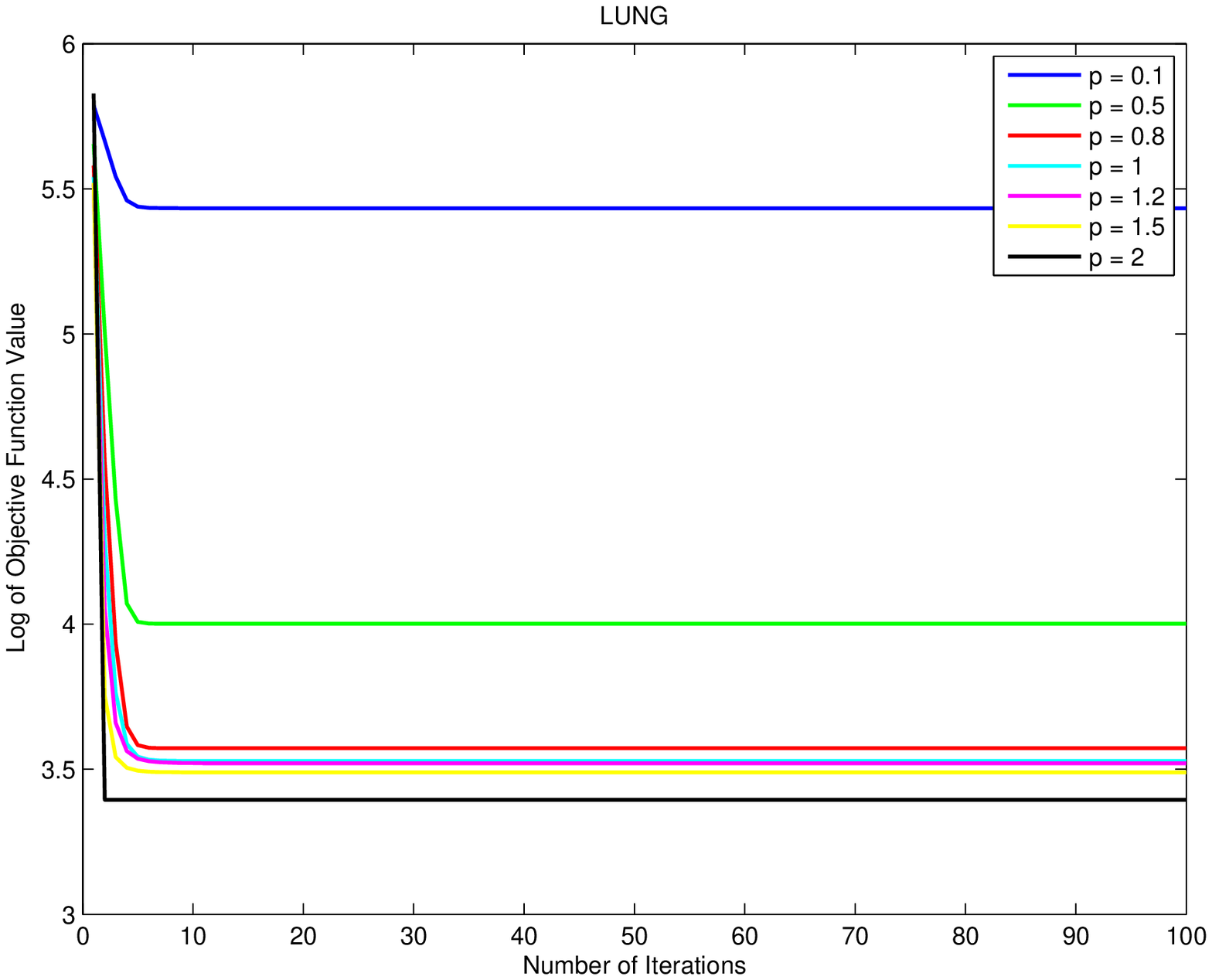}}
  \subfigure[ALLAML]{
    \includegraphics[width=0.23\textwidth]{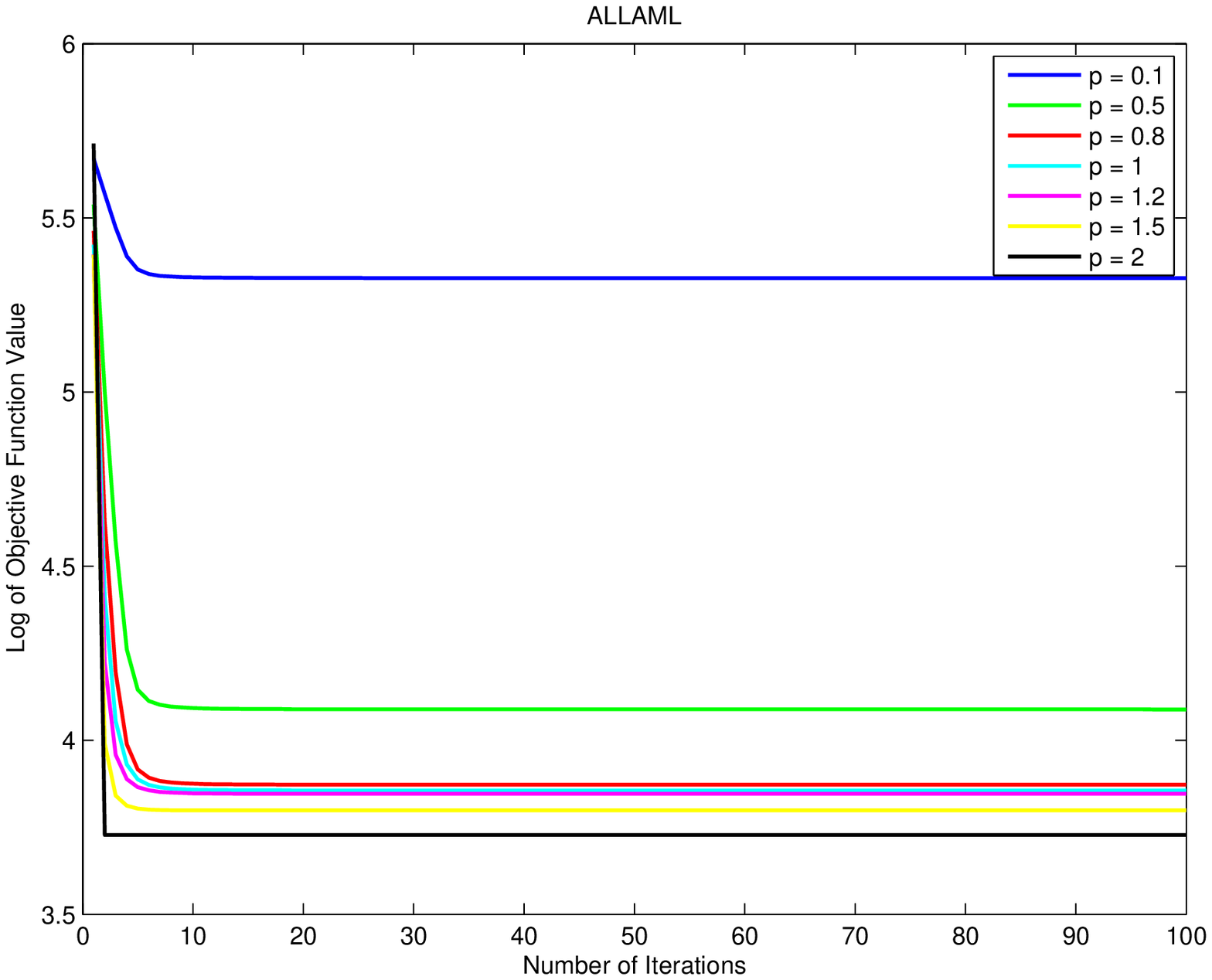}}
  \subfigure[TOX−171]{
    \includegraphics[width=0.23\textwidth]{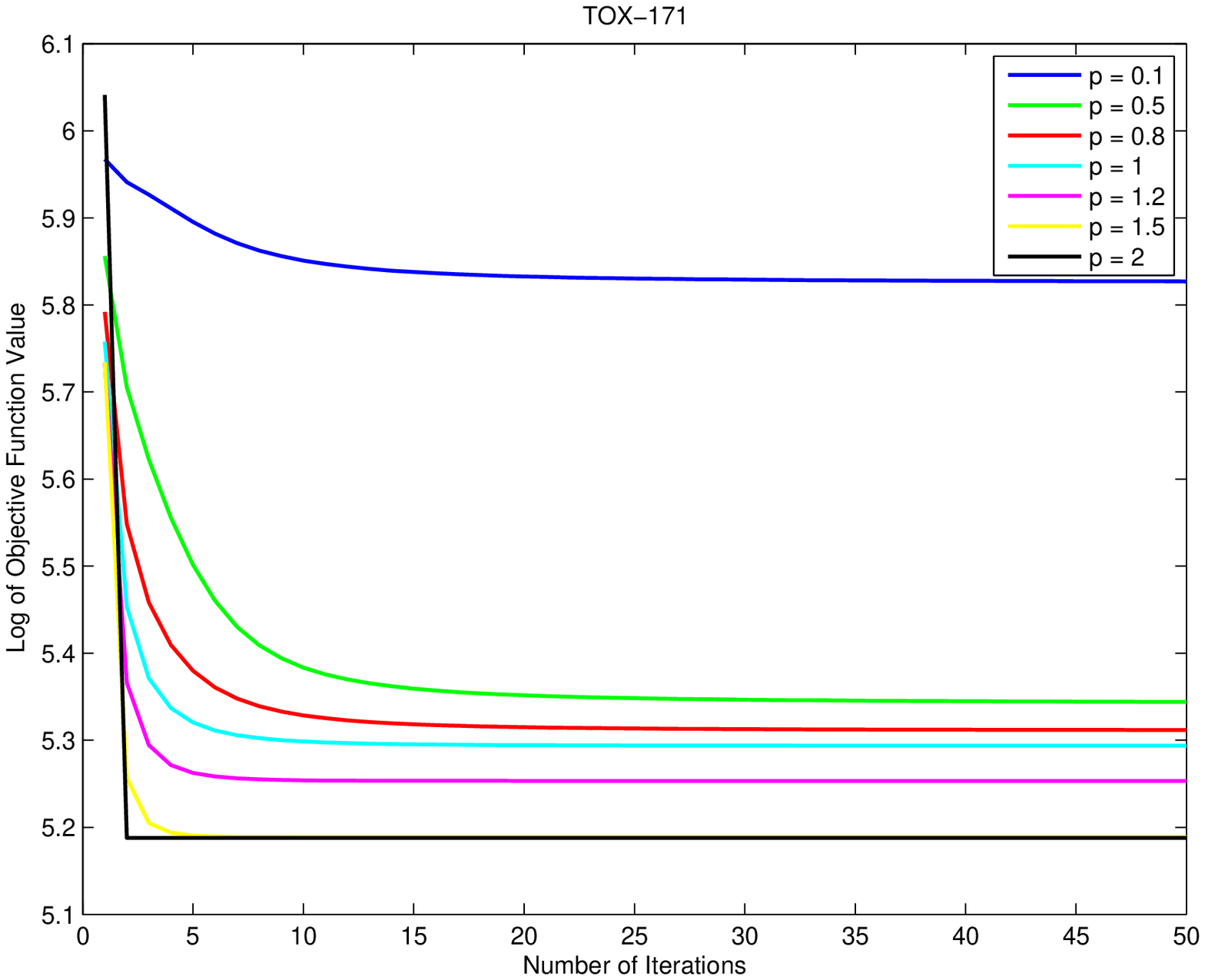}}
  \subfigure[AR10P]{
    \includegraphics[width=0.23\textwidth]{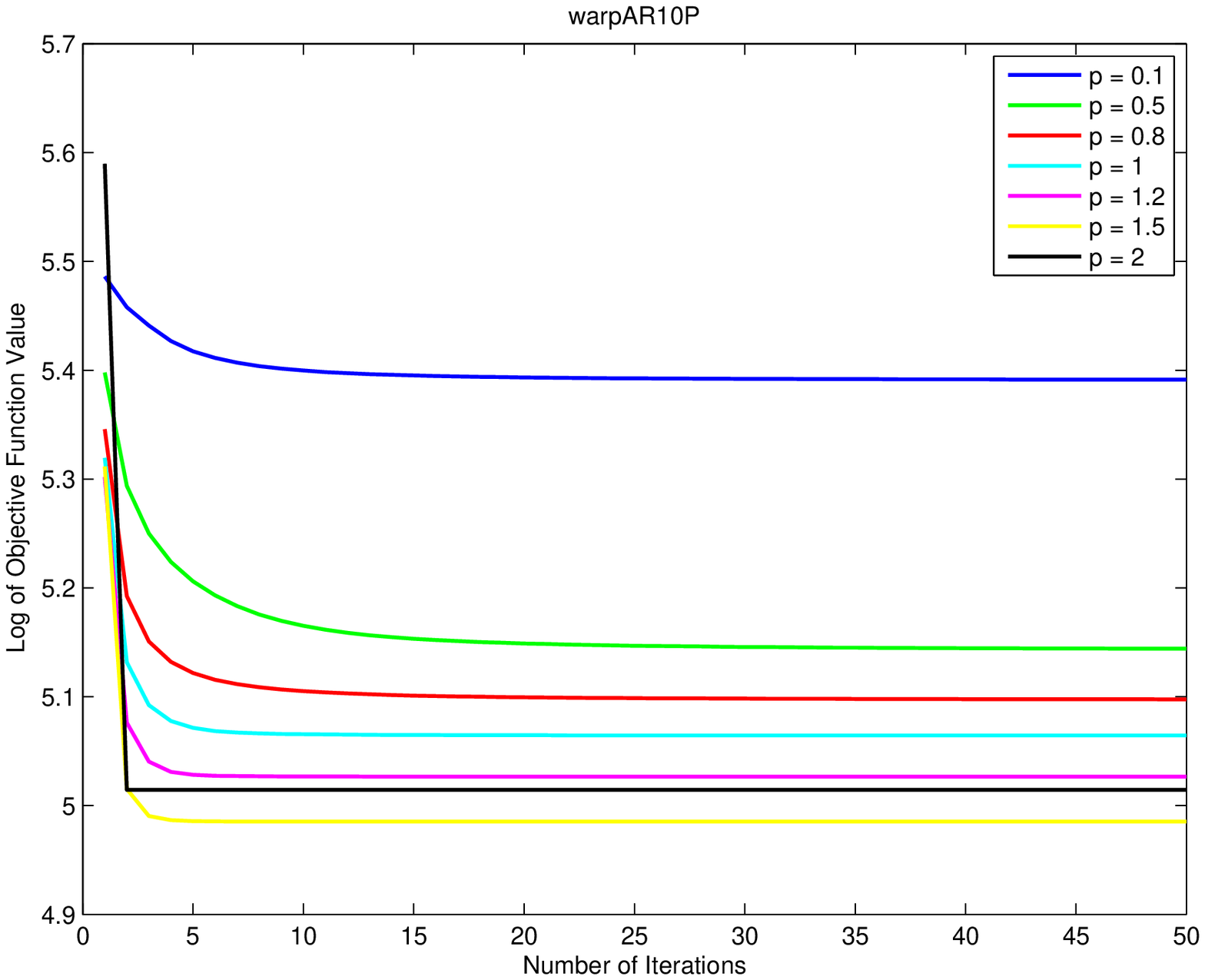}}
\caption{Log of objective function value with different $p$ value. The objective function is Eq.~\eqref{ex2}.}
\end{figure*}

\subsection{Experiments on a Robust Feature Selection Problem}
\begin{figure*}[!t]
\label{fig3} 
  \centering
  \subfigure[DBWorld]{
    \includegraphics[width=0.23\textwidth]{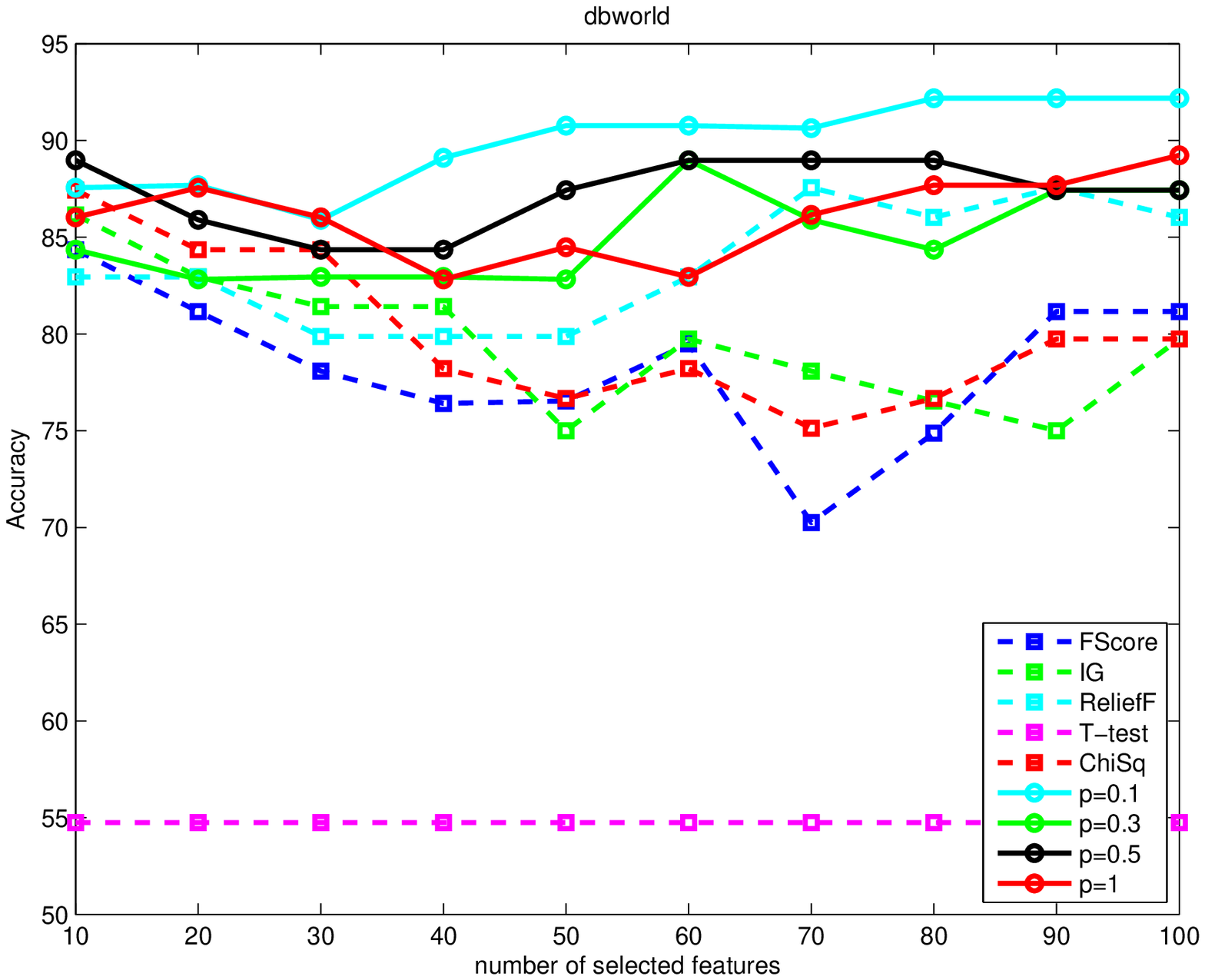}}
  \subfigure[Prostate−GE]{
    \includegraphics[width=0.23\textwidth]{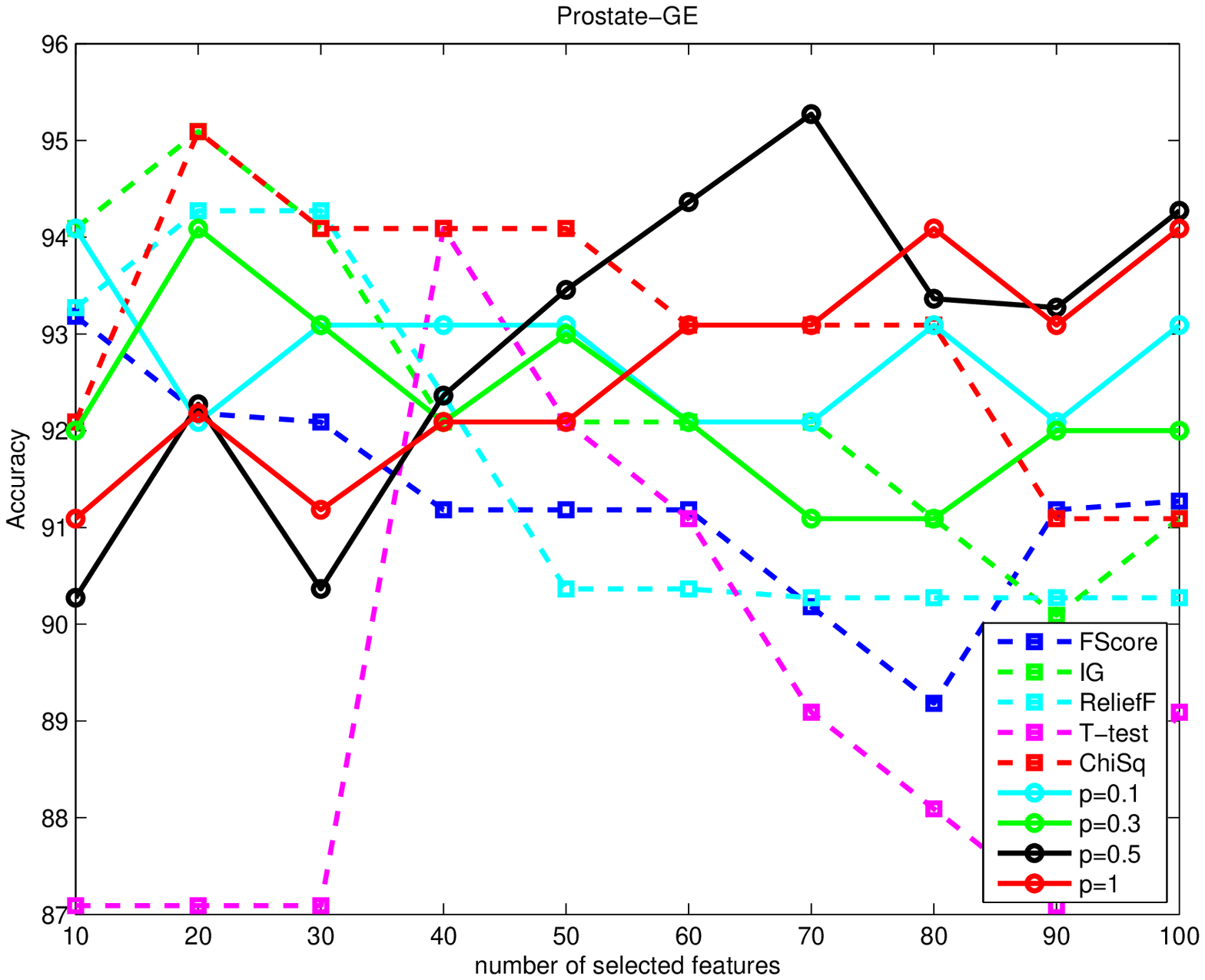}}
  \subfigure[Carcinomas]{
    \includegraphics[width=0.23\textwidth]{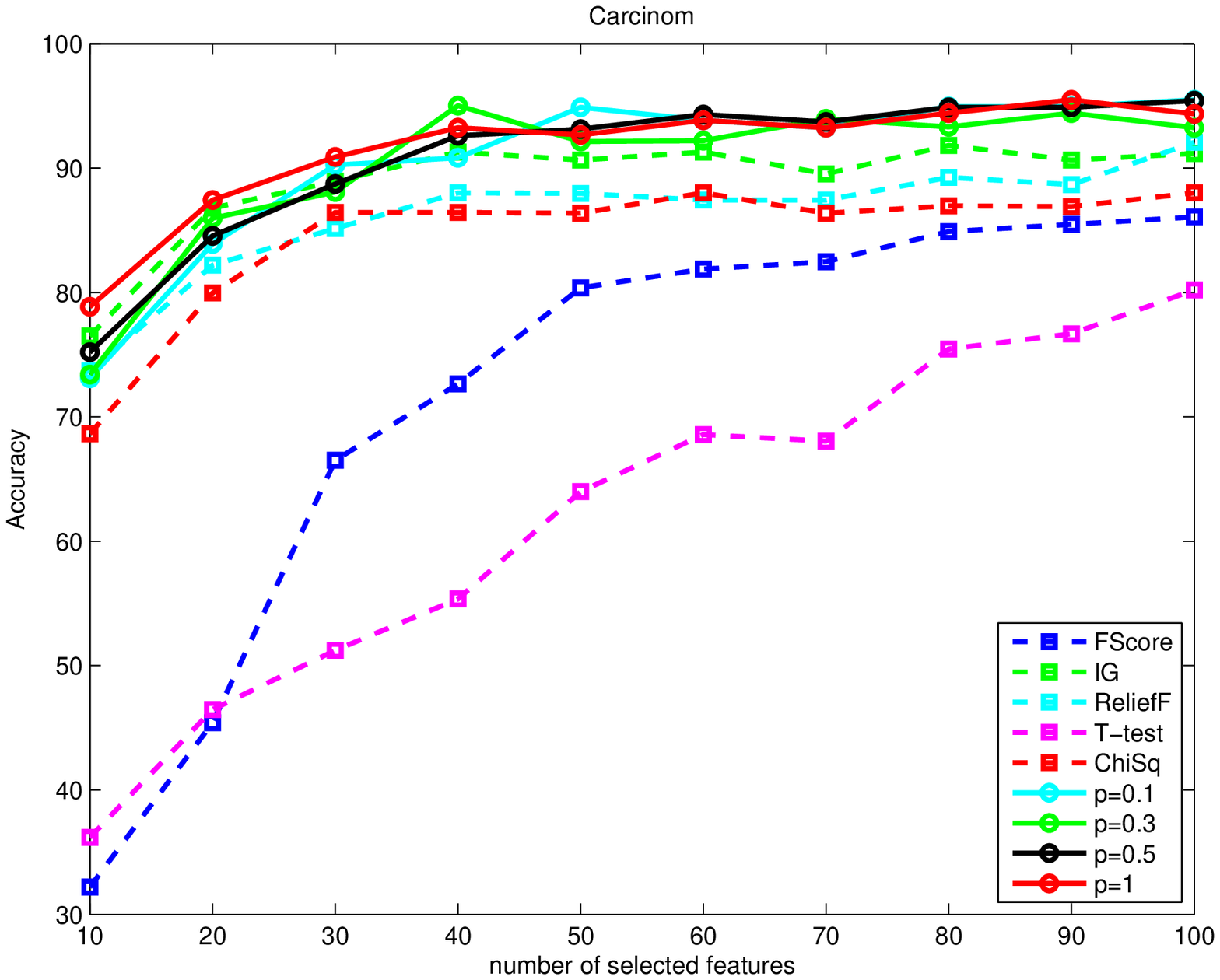}}
  \subfigure[GLIOMA]{
    \includegraphics[width=0.23\textwidth]{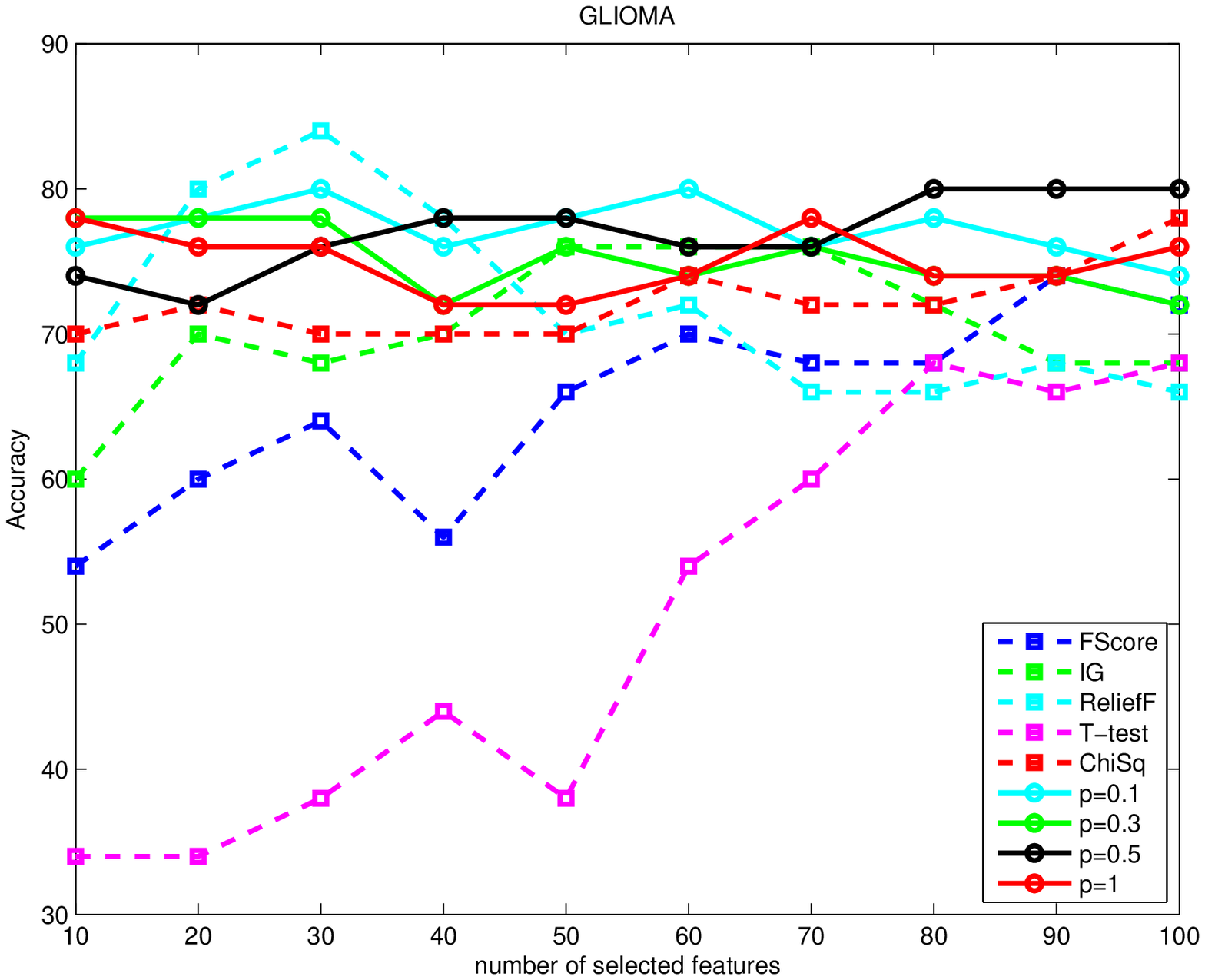}}
  \subfigure[LUNG]{
    \includegraphics[width=0.23\textwidth]{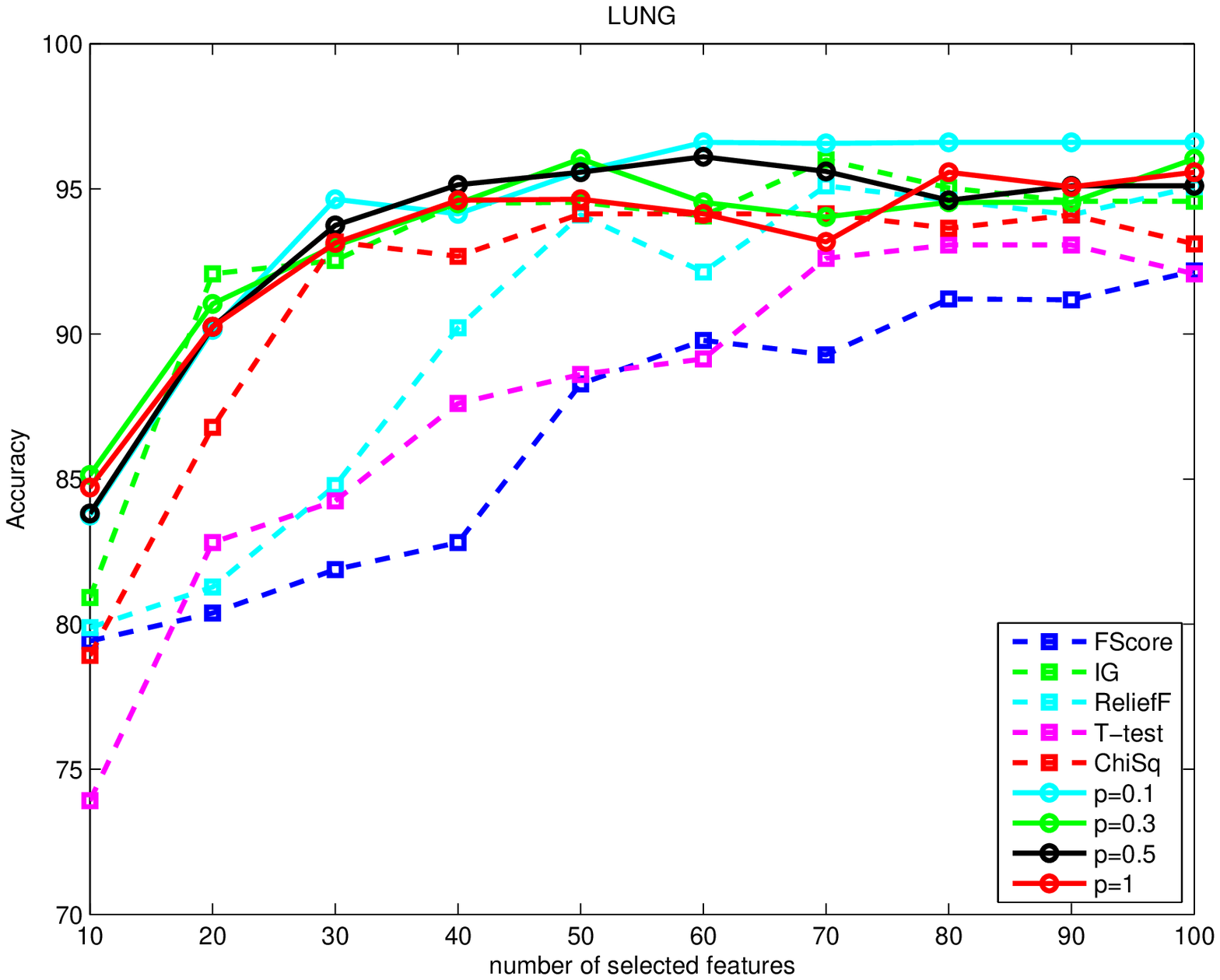}}
  \subfigure[ALLAML]{
    \includegraphics[width=0.23\textwidth]{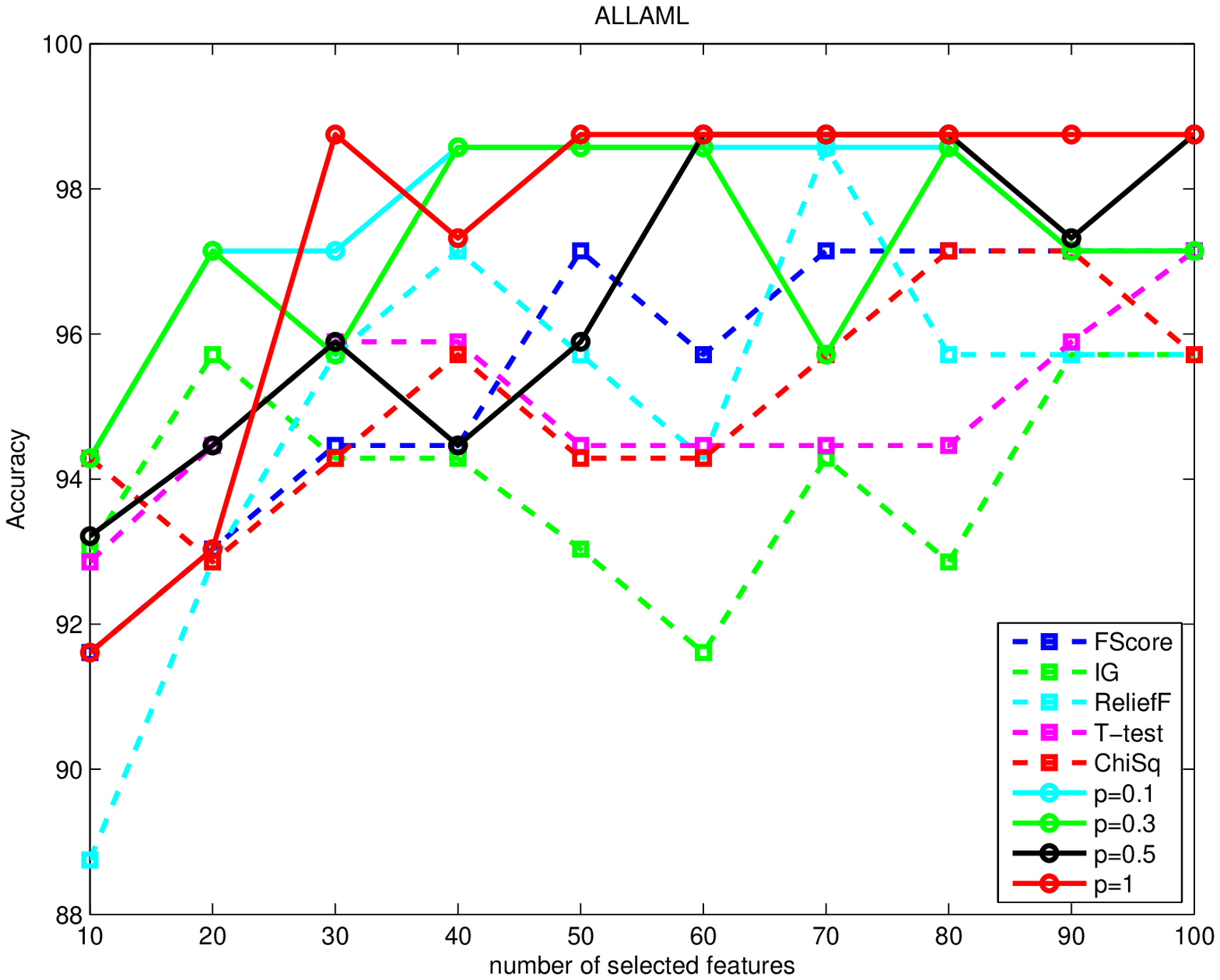}}
  \subfigure[SMK−CAN−187]{
    \includegraphics[width=0.23\textwidth]{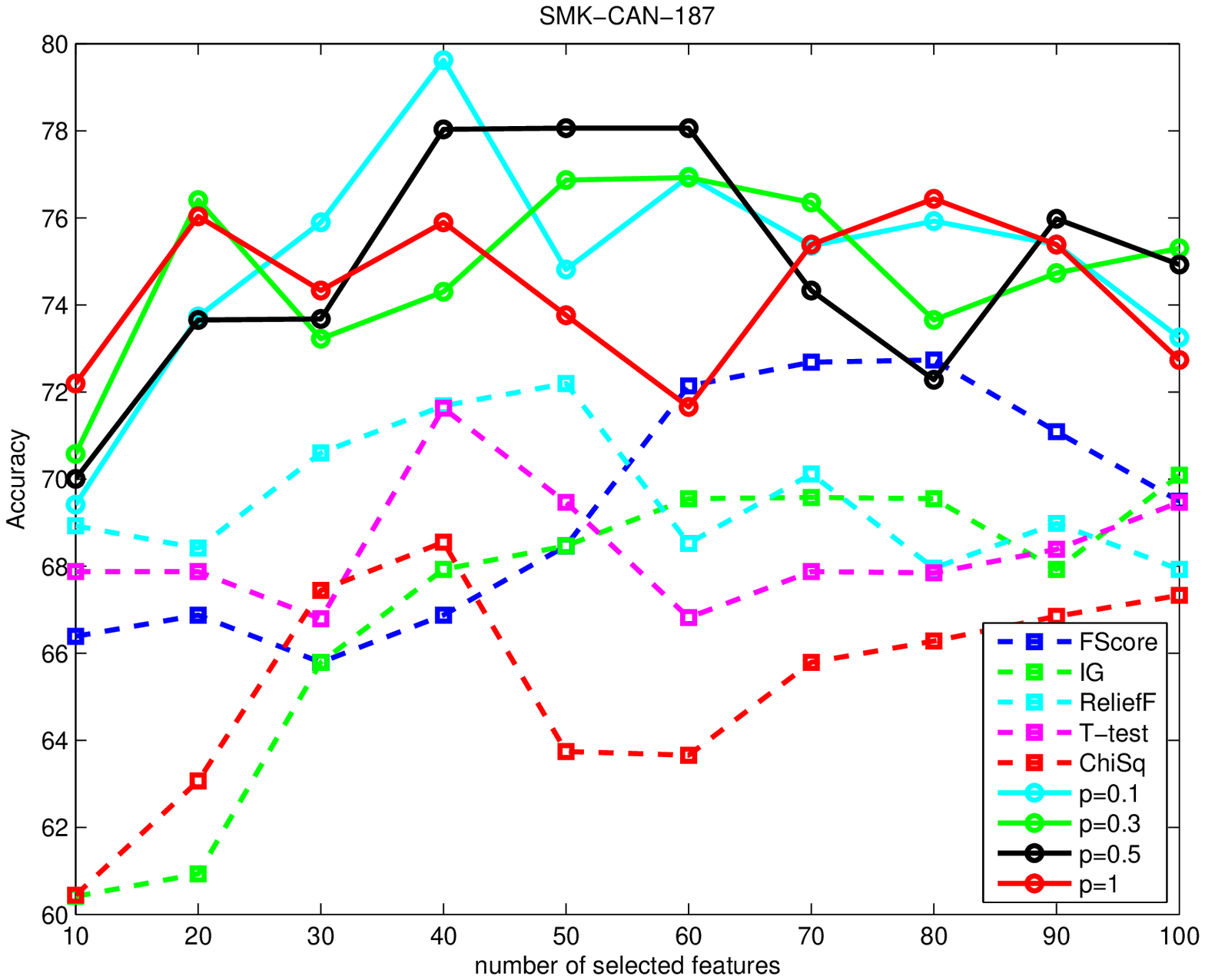}}
  \subfigure[TOX−171]{
    \includegraphics[width=0.23\textwidth]{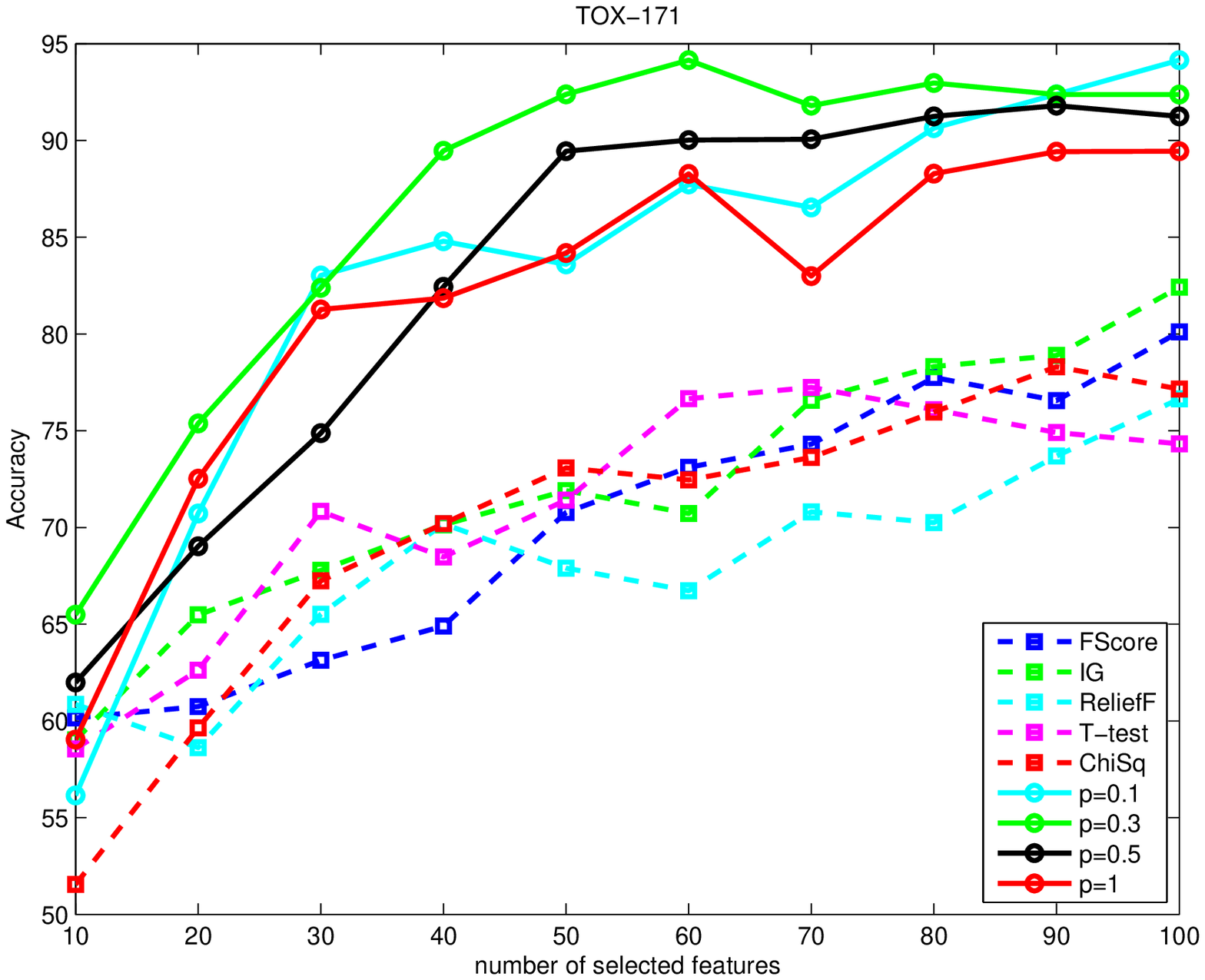}}
  \subfigure[AR10P]{
    \includegraphics[width=0.23\textwidth]{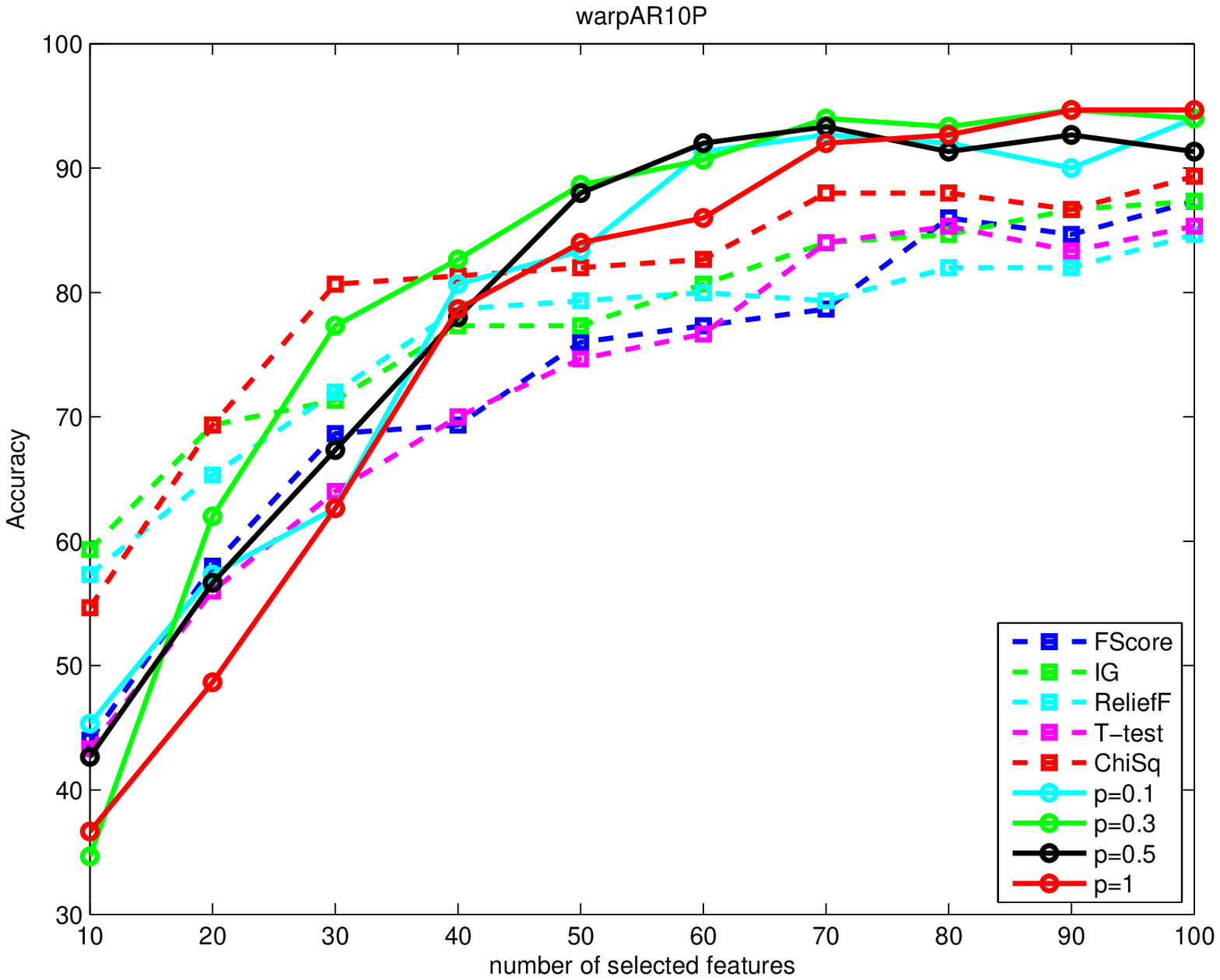}}
  \subfigure[PIE10P]{
    \includegraphics[width=0.23\textwidth]{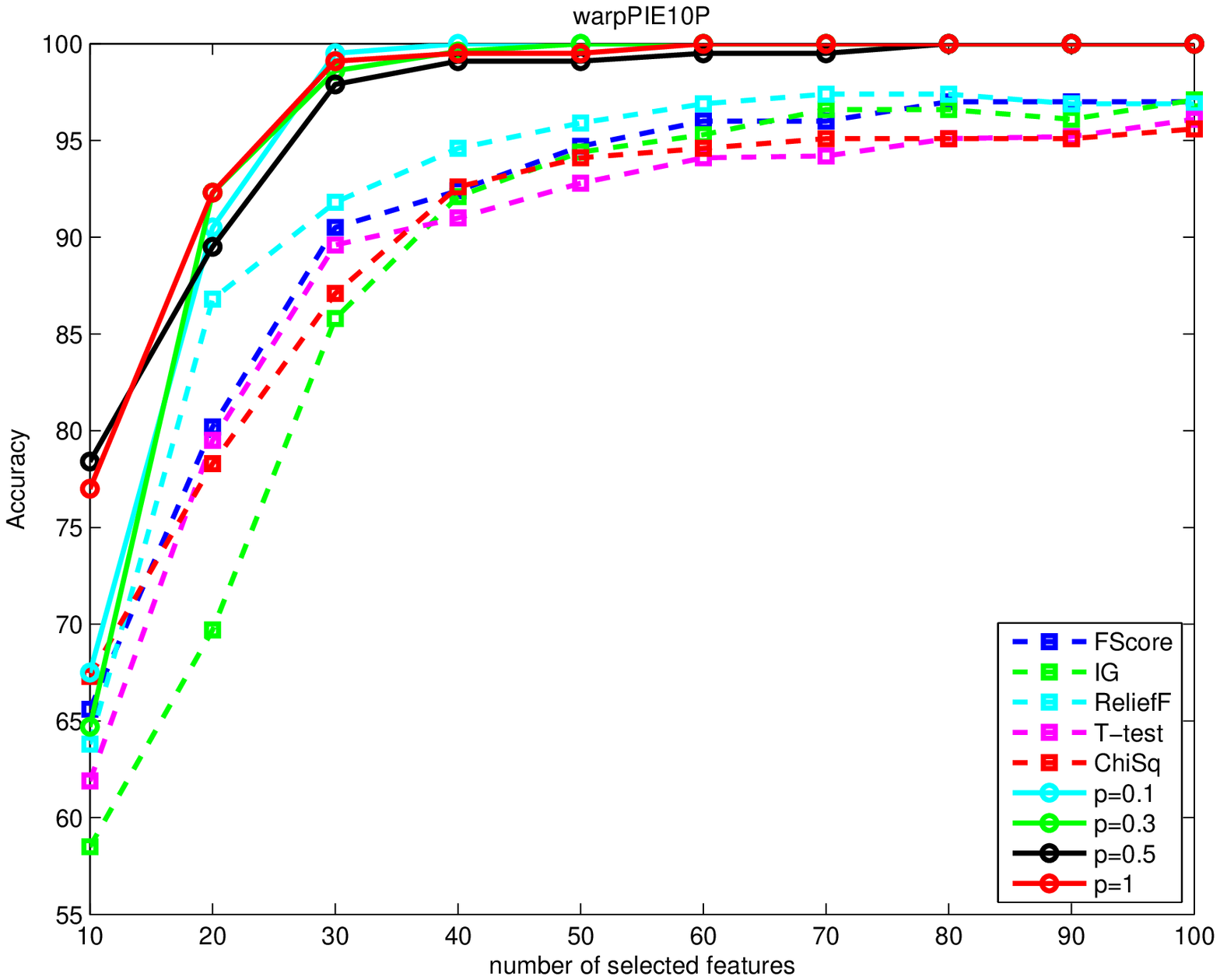}}
  \subfigure[Arcene]{
    \includegraphics[width=0.23\textwidth]{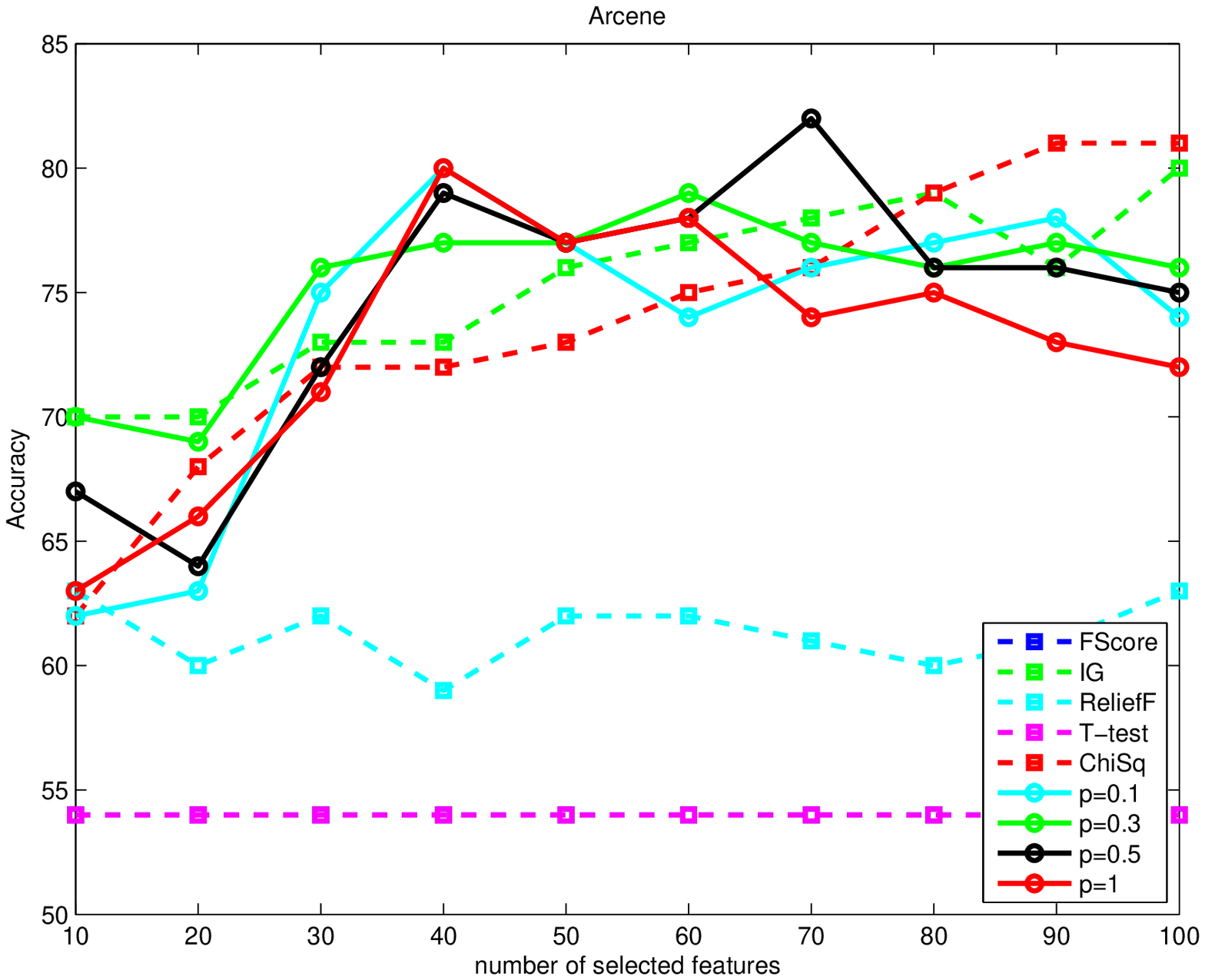}}
  \subfigure[CLL−SUB−111]{
    \includegraphics[width=0.23\textwidth]{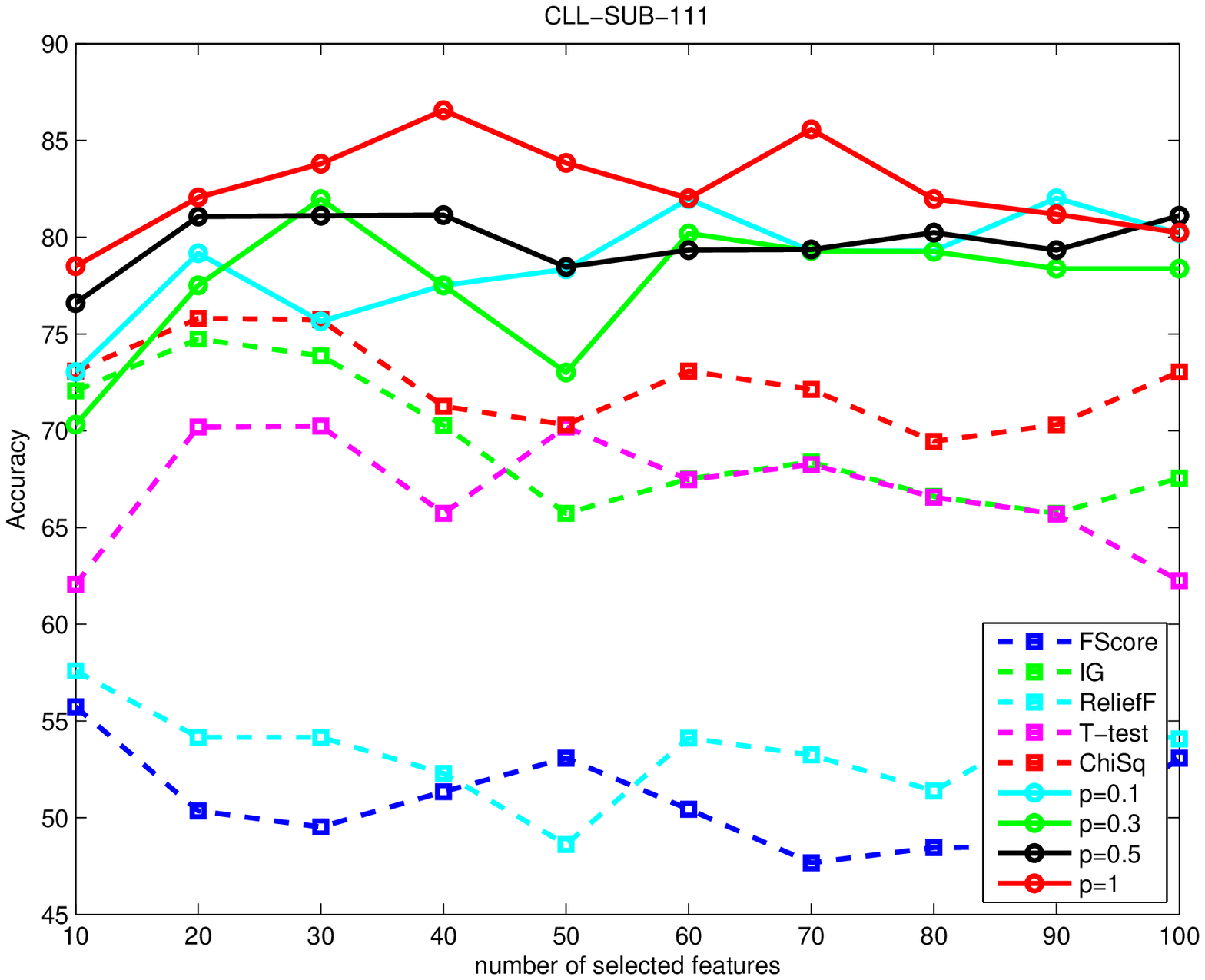}}
     \subfigure[MLLML]{
    \includegraphics[width=0.23\textwidth]{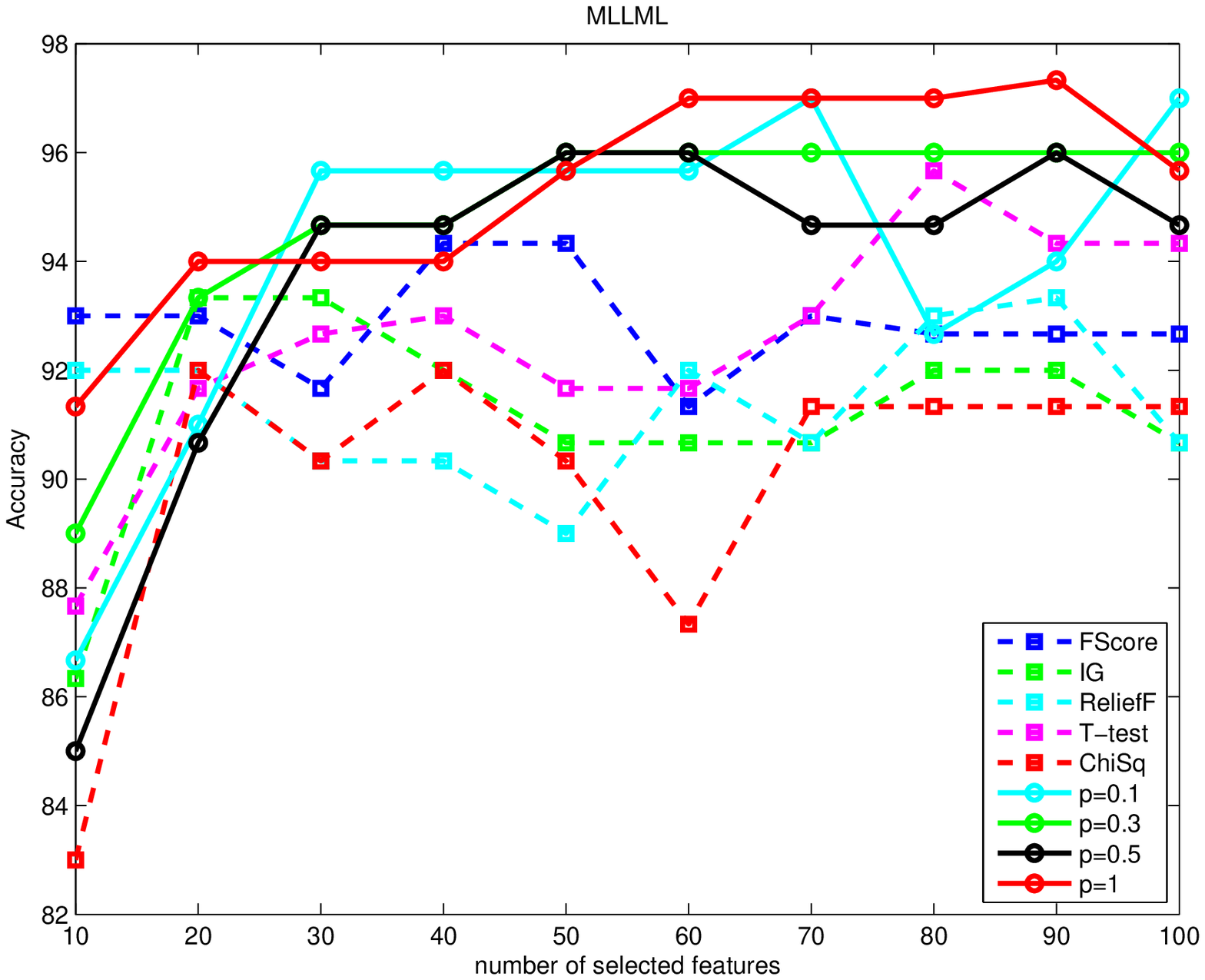}}
  \subfigure[ORL10P]{
    \includegraphics[width=0.23\textwidth]{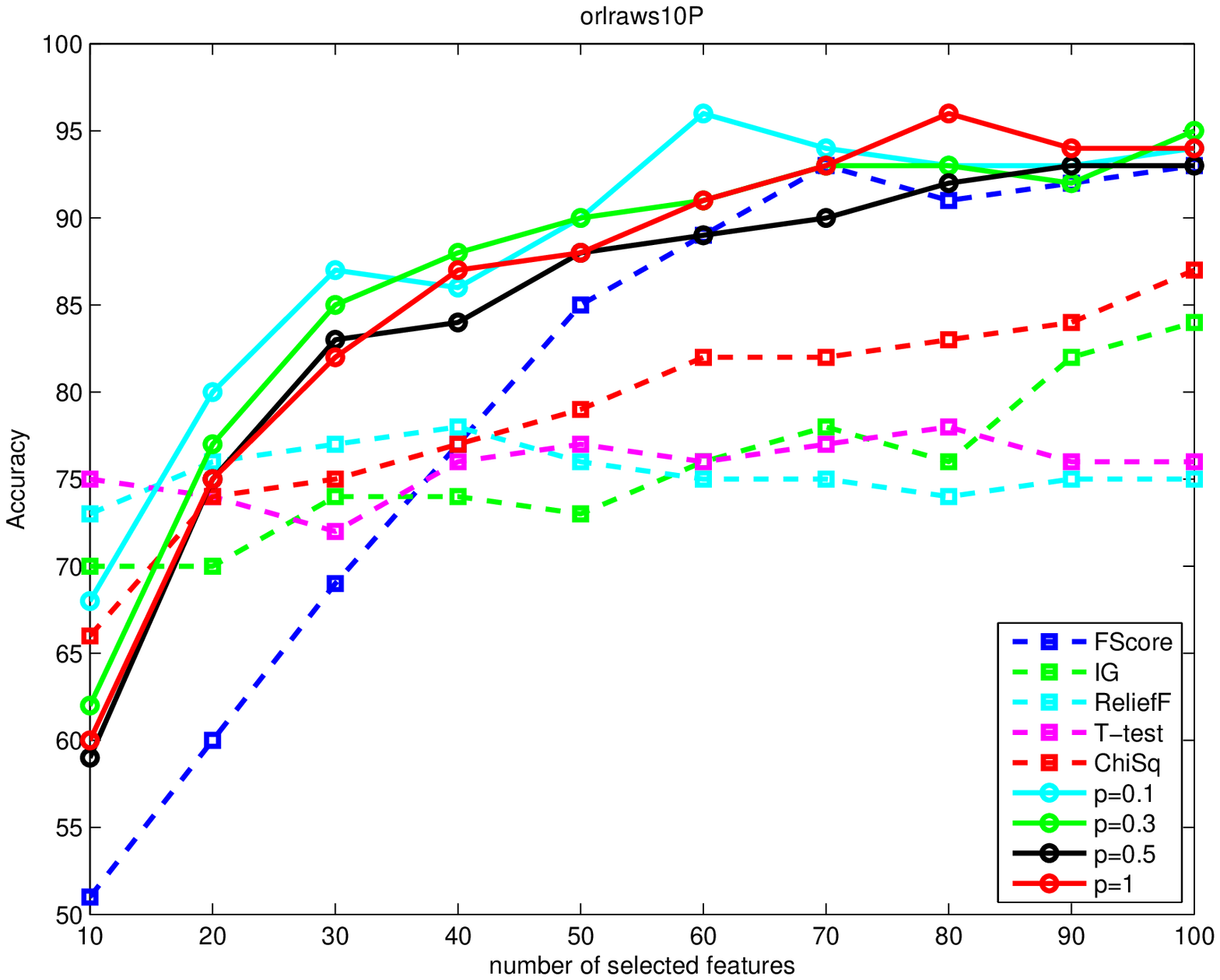}}
  \subfigure[PIX10P]{
    \includegraphics[width=0.23\textwidth]{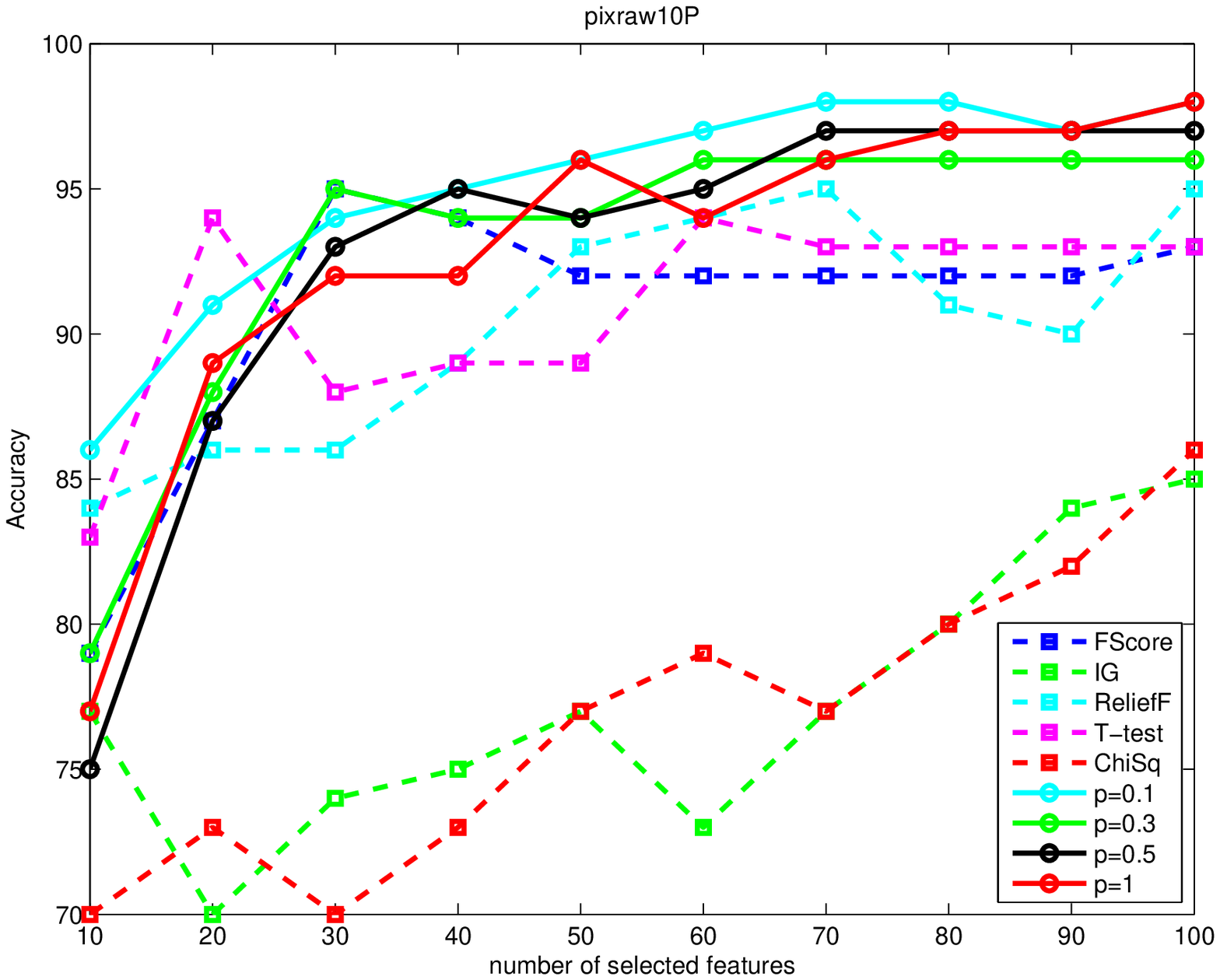}}
  \subfigure[Prostate−MS]{
    \includegraphics[width=0.23\textwidth]{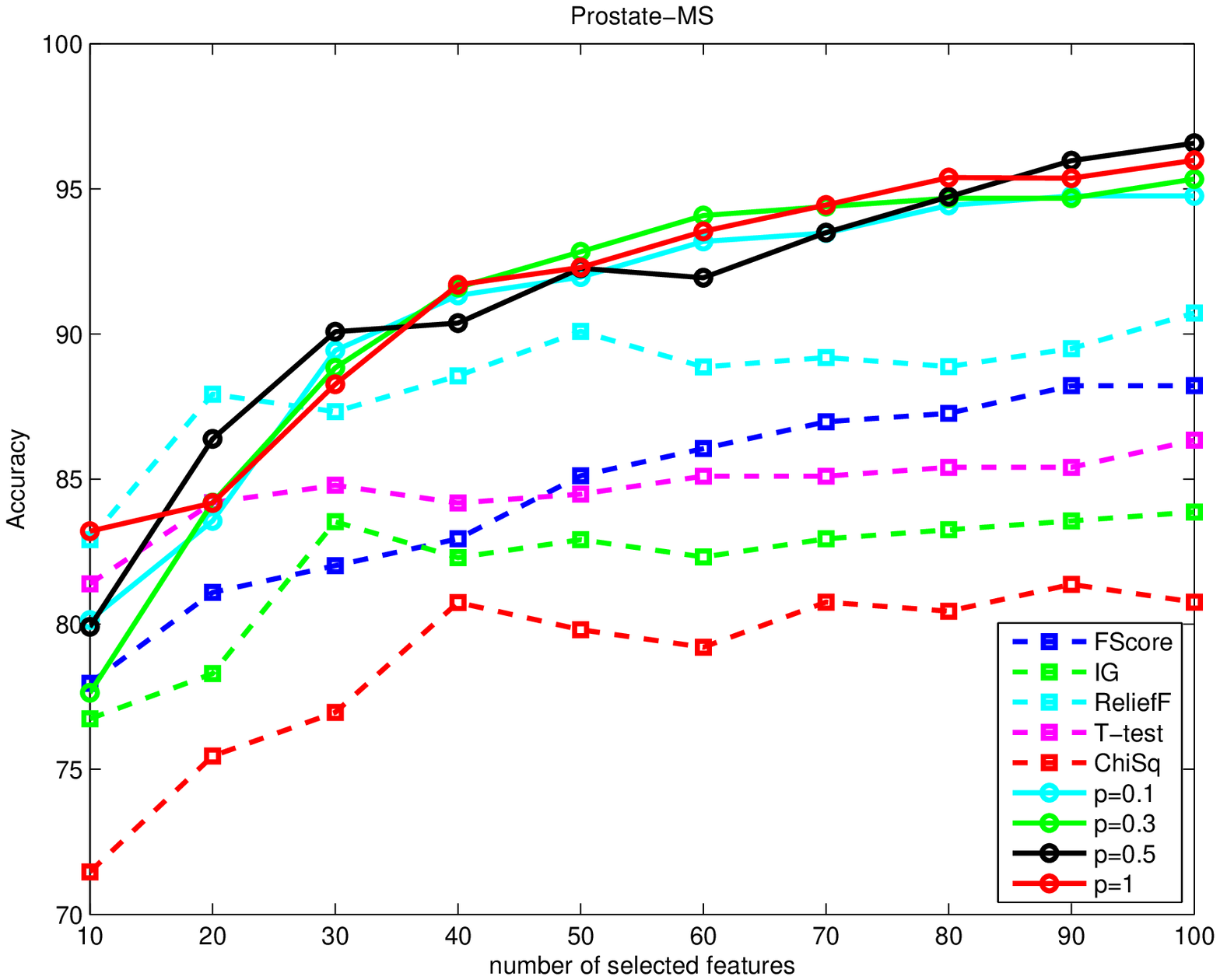}}
\caption{Classification Accuracy of different methods on selected datasets.}
\vbox{}
\end{figure*}
In this subsection, we applied our method to a robust feature selection problem. We utilized our algorithm to solve the following question so as to find an appropriate weight matrix $W$ with which we can accomplish efficient feature selection.
\begin{equation}
\label{ex30}
\mathop {\min }\limits_{M \ge 0,W} {\left\| {{X^T}W + 1{b^T} - Y - Y \circ M} \right\|_{2,1}} + \gamma \left\| W \right\|_{2,p}^p
\end{equation}

In the above function, $\circ$ is the Hadamard product, and M is defined as $M = max(({X^T}W + 1{b^T} - Y) \circ Y,~0)$. By means of the matrix $M$, we implemented a classifier where all positive loss brought by those correctly classified points was discarded. This trick brought more robustness to our method than other related ridge regression classification methods since a correctly classified points won't generate loss in the objective.

Before showing experimental results, we first briefly summarize the solving process of Problem (\ref{ex30}). Resorting to Algorithm \ref{alg1}, Problem (\ref{ex30}) can be rewritten as an easily solvable form as below:
\begin{small}
\begin{equation}
\label{ex3}
\begin{split}
\mathop {\min }\limits_{M \ge 0,W} &{tr(({X^T}W + 1{b^T}-Y-Y \circ M)^TD_1({X^T}W + 1{b^T}-Y-Y \circ M))}\\
+ &\gamma tr(W^TD_2W)\,.
\end{split}
\end{equation}
\end{small}
where $D_1$ is a diagonal matrix with the $k$-th diagonal element to be $\frac{1}{2}(({X^T}w^k + ({b^T} - y^k - y^k \circ m^k))^T({X^T}w^k + ({b^T} - y^k - y^k \circ m^k))+\delta)^{-\frac{1}{2}}$ and $D_2$ is a diagonal matrix with the $k$-th diagonal element to be $\frac{p}{2}((w^k)^T w^k + \delta)^{\frac{p-2}{2}}$.

Similarly, we can solve problem (\ref{ex3}) via the alternative optimization method. Taking derivative w.r.t. $w_i$ in problem (\ref{ex3}) and we get:
\begin{equation}
\mathop XD_1X^TW + XD_1(1{b^T} - Y - Y \circ M)+ \gamma D_2W = 0\,.
\end{equation}
That is,
\begin{equation}
\mathop W = (XD_1X^T + \gamma D_2)^{-1}(XD_1(Y + Y \circ M - 1{b^T})\,.
\end{equation}
Taking derivative w.r.t. $b$ in problem (\ref{ex3}) and we get:
\begin{equation}
\mathop ({X^T}W - Y - Y \circ M)^TD_21 + b1^TD_21 = 0\,.
\end{equation}
The optimal solution of $b$ is:
\begin{equation}
b = \frac{(-{X^T}W +Y + Y \circ M)^TD_21}{1^TD_21}\,.
\end{equation}

For this problem, we update variables $W$, $D$, $b$ and $M$ alternatively and iteratively until convergence.

In this test, We applied all $16$ data sets to this experiment. For each data set evaluated in the experiment, we employed the 5-fold cross validation, which randomly selected 80\% of the data for training and the remaining $20\%$ for testing. We utilized the SVM classifier with linear kernel and let $C = 1$. The number of features selected ranges from 10 to 100, with the incremental step to be 10. We compared our proposed feature selection method with several popularly used feature selection methods: Fisher Score \cite{duda1973pattern}, Information Gain \cite{Raileanu00theoreticalcomparison}, ReliefF \cite{Relief92, ReliefFKononenko1994}, T-test and ChiSquare \cite{Manning2008}.

For our method, we collected the performance for four different $p$ values, which are $\{0.1, 0.3, 0.5, 1\}$. We didn't use $p$ values larger than $1$ since we want to guarantee the sparsity of the $W$ matrix. Also, as we have validated in the previous two experiments, our method converges very fast, hence we set the number of iteration times to be $30$.
The evaluation of different methods is based on the average classification accuracy, which is summarized in Fig.~3 and Fig.~4.

Observing the Fig.~3 and Fig.~4, we are confirmed with the effectiveness of our proposed method on real benchmark data sets. Our method generally has a high potential to outperform other traditional methods on these various kinds of data sets. No matter what the $p$ value is, our method always gains its superiority. Also, the efficiency of our method has been discussed and demonstrated previously. All in all, our algorithm is capable of finding a promising classification matrix, which is more robust to outliers and finishes with guaranteed speed.

\section{Conclusions}
Loss function and regularizer are two significant factors influencing the performance of an algorithm. And, for each of them the Sparsity-Inducing Norms is generally involved. In order to solve the complicated problems with Sparsity-Inducing Norms, in this work we provide a simple yet efficient optimization method, which can cope with the case that both loss function and regularizer are non-convex. The proposed method is suitable for various tasks.

Two  issues of IRW are:1) theoretically, only stationary points are provided for non-convex problems; 2) practically, IRW is not efficient for problems with multiple inseparable variables, for the closer-form solution cannot be directly obtained when the surrogate function has multiple inseparable variables. Solving these two issues is the focus in our future work.


%
%
%

\vskip 0.2in

\section*{Acknowledgment}

This work was supported in part by the National Natural
Science Foundation of China grant under numbers 61772427 and
61751202.

\ifCLASSOPTIONcaptionsoff
  \newpage
\fi



\bibliographystyle{IEEEtran}
\bibliography{elsart123}
\end{document}